%% file: main.tex
\documentclass[10pt,twocolumn,letterpaper]{article}

\usepackage[pagenumbers]{cvpr} 

\input{preamble}

\definecolor{cvprblue}{rgb}{0.21,0.49,0.74}
\usepackage[pagebackref,breaklinks,colorlinks,allcolors=cvprblue]{hyperref}
\usepackage{multirow}
\usepackage{tabu}
\usepackage{colortbl}
\usepackage{makecell}
\usepackage{booktabs}
\usepackage{pifont}
\usepackage{bbm}
\usepackage{graphicx}
\usepackage{algorithm}
\usepackage{algpseudocode}
\usepackage{subcaption}
\usepackage{amsfonts}
\usepackage{siunitx}

\def\paperID{****} 
\def\confName{CVPR}
\def\confYear{2025}

\title{AR4D: Autoregressive 4D Generation from Monocular Videos}

\author{
    \textbf{Hanxin Zhu$^{1}$\thanks{This paper is the result of an open source project starting from March 2024.}\;, ~Tianyu He$^{2}$, ~Xiqian Yu$^{1}$, ~Junliang Guo$^{2}$, ~Zhibo Chen$^{1}$, ~Jiang Bian$^{2}$} \\
    $^{1}$University of Science and Technology of China \:
    $^{2}$Microsoft Research Asia \\
    \texttt {\{hanxinzhu, yuxiqian\}@mail.ustc.edu.cn, chenzhibo@ustc.edu.cn} \\
    \texttt {\{tianyuhe, junliangguo, jiang.bian\}@microsoft.com} \\
    ~\textbf{\normalsize{\url{https://hanxinzhu-lab.github.io/AR4D/}}
   }
}

\begin{document}
\maketitle

\input{sec/0_abstract}    
\input{sec/1_intro}
\input{sec/2_related}
\input{sec/3_preliminaries}

\input{sec/4_method}
\input{sec/5_experiments}
\input{sec/6_discussion}

{
    \small
    \bibliographystyle{ieeenat_fullname}
    \bibliography{main}
}


\end{document}

%% file: preamble.tex
\usepackage[dvipsnames]{xcolor}
\usepackage{calc}

\definecolor{tabfirst}{rgb}{1, 0.7, 0.7}
\definecolor{tabsecond}{rgb}{1, 0.85, 0.7}
\definecolor{tabthird}{rgb}{1, 1, 0.7}
\definecolor{tabgray}{rgb}{0.9, 0.9, 0.9}
\definecolor{turquoise}{cmyk}{0.65,0,0.1,0.3}
\definecolor{purple}{rgb}{0.65,0,0.65}
\definecolor{darkgreen}{rgb}{0, 0.5, 0}
\definecolor{orange}{rgb}{0.8, 0.6, 0.2}
\definecolor{red}{rgb}{0.8, 0.2, 0.2}
\definecolor{darkred}{rgb}{0.6, 0.1, 0.05}
\definecolor{blueish}{rgb}{0.0, 0.3, .6}
\definecolor{light_gray}{rgb}{0.7, 0.7, .7}
\definecolor{pink}{rgb}{1, 0, 1}
\definecolor{greyblue}{rgb}{0.25, 0.25, 1}

\def\eg{\emph{e.g}\onedot}

\def\ie{\emph{i.e}\onedot}

\newcommand\rgt{\aftergroup\mathclose\aftergroup{\aftergroup}\right}


%% file: sec/0_abstract.tex
\begin{abstract}
Recent advancements in generative models have ignited substantial interest in dynamic 3D content creation (\ie, 4D generation). Existing approaches primarily rely on Score Distillation Sampling (SDS) to infer novel-view videos, typically leading to issues such as limited diversity, spatial-temporal inconsistency and poor prompt alignment, due to the inherent randomness of SDS. To tackle these problems, we propose AR4D, a novel paradigm for SDS-free 4D generation. Specifically, our paradigm consists of three stages. To begin with, for a monocular video that is either generated or captured, we first utilize pre-trained expert models to create a 3D representation of the first frame, which is further fine-tuned to serve as the canonical space. Subsequently, motivated by the fact that videos happen naturally in an autoregressive manner, we propose to generate each frame's 3D representation based on its previous frame's representation, as this autoregressive generation manner can facilitate  more accurate geometry and motion estimation. Meanwhile, to prevent overfitting during this process, we introduce a progressive view sampling strategy, utilizing priors from pre-trained large-scale 3D reconstruction models. To avoid appearance drift introduced by autoregressive generation, we further incorporate a refinement stage based on a global deformation field and the geometry of each frame’s 3D representation. Extensive experiments have demonstrated that AR4D can achieve state-of-the-art 4D generation without SDS, delivering greater diversity, improved spatial-temporal consistency, and better alignment with input prompts. 
\end{abstract}

%% file: sec/1_intro.tex
\section{Introduction}
\label{sec:intro}
\begin{figure}[t]
    \centering
    \includegraphics[width=1\linewidth]{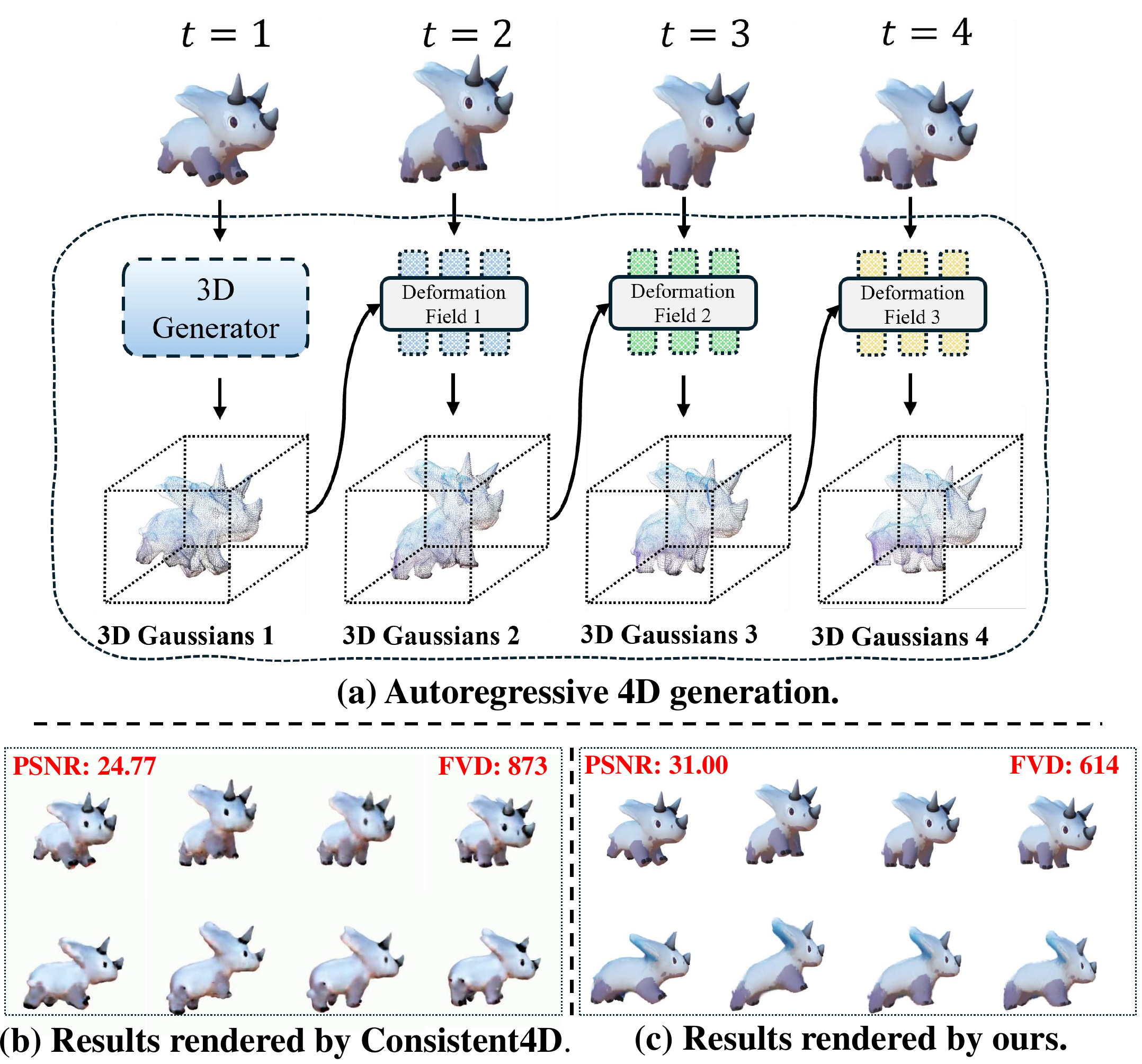}
    \caption{\textbf{Illustration of autoregressive 4D generation.} In comparison to SDS-based methods (\eg, Consistent4D~\cite{jiang2023consistent4d}), our approach enables SDS-free 4D generation with substantial advancements, including better alignment with input videos and improved spatial-temporal consistency, etc.}
    \label{fig:teaser}
\end{figure}

In recent years, generative models have made significant strides, allowing for the generation of highly realistic images~\cite{rombach2022high,zhang2023adding,mou2024t2i,podell2023sdxl} and videos~\cite{wu2023tune,blattmann2023stable,villegas2022phenaki,zhang2024show} from simple prompts. Building on these successes, numerous studies have sought to extend these capabilities into the domain of dynamic 3D content creation (\ie, 4D generation)~\cite{jiang2023consistent4d, ren2023dreamgaussian4d, zhao2023animate124,sun2024eg4d,yang2024diffusion}, which is crucial for areas such as virtual reality, gaming, and embodied intelligence.

To achieve this goal, given the lack of large-scale 4D datasets available, existing methods~\cite{jiang2023consistent4d,ling2024align,bahmani20244d,ren2023dreamgaussian4d,zeng2025stag4d,bahmani2025tc4d,gao2024gaussianflow,miao2024pla4d,jiang2024animate3d,yuan20244dynamic,li2024dreammesh4d,zhao2023animate124,zhu2024compositional} mainly estimate novel-view videos using Score Distillation Sampling (SDS)~\cite{poole2022dreamfusion}, where knowledge stored in pre-trained  multi-modal diffusion models~\cite{liu2023zero,liu2024one,blattmann2023stable} are leveraged to guide the generation process.  However, while seemingly reasonable results can be obtained, these SDS-based methods often exhibit several issues~\cite{wang2024prolificdreamer,liang2024luciddreamer,yi2023gaussiandreamer}, \eg, limited diversity, spatial-temporal inconsistencies, poor alignment with input prompts, typically resulting in low-quality 4D objects, as demonstrated in Fig.~\ref{fig:teaser}(b).

\begin{figure*}[t]
    \centering
\includegraphics[width=1\linewidth]{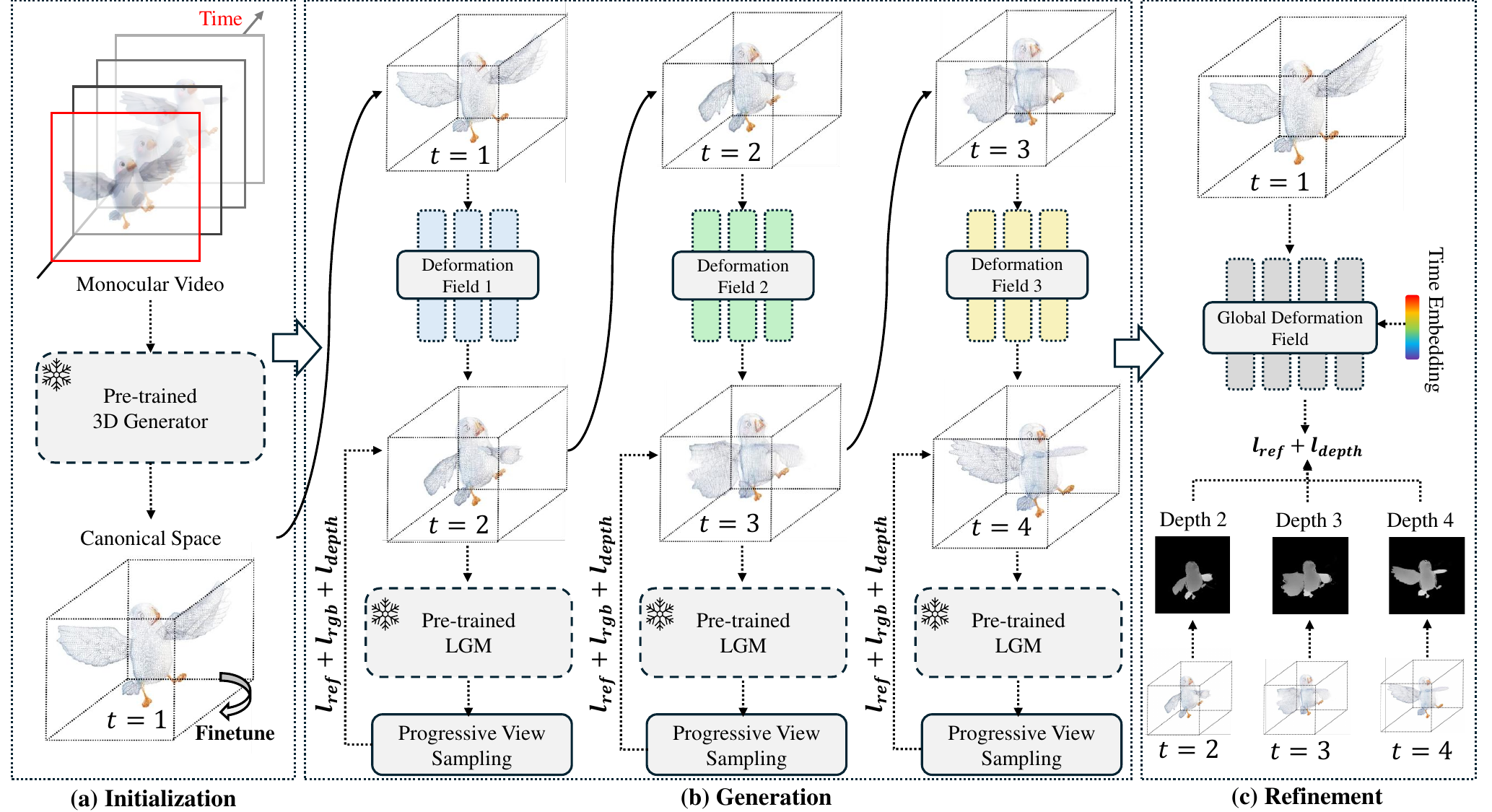}
    \caption{\textbf{Paradigm of our proposed AR4D.} To enable SDS-free 4D generation, we propose a three-stage approach consisting of \textbf{\textit{Initialization}}, \textbf{\textit{Generation}}, and \textbf{\textit{Refinement}}. Please see Sec.~\ref{sec:methods} for more details.}
    \label{fig:framework}
\end{figure*}

To address these issues, in this paper we propose AR4D, a novel paradigm capable of generating high-quality 4D assets without relying on SDS. Specifically, as shown in Fig.~\ref{fig:framework}, our paradigm is composed of three distinct stages, which are refered to as the \textbf{\textit{Initialization}} stage, the \textbf{\textit{Generation}} stage, and the \textbf{\textit{Refinement}} stage respectively. To begin with, during the \textbf{\textit{Initialization}} stage, as shown in Fig.~\ref{fig:teaser}(a), given a monocular video (either generated or captured), we first utilize pre-trained 3D generators (\eg, MVDream~\cite{shi2023mvdream}, Splatt3R~\cite{smart2024splatt3r}, etc.) to create a 3D representation (\ie, 3D Gaussians~\cite{kerbl20233d}) of the first frame, which is further fine-tuned to serves as the canonical space for the 4D content to be generated. 

Subsequently, during the \textbf{\textit{Generation}} stage, to derive the corresponding 4D asset based on the reference video and its first frame's 3D representation without relying on SDS, an intuitive way is to directly employ established 4D reconstruction methods , \eg, Deform 3DGS~\cite{yang2024deformable}, which learns the deformation of the canonical space through a global deformation field by minimizing the difference between rendered and ground-truth frames. However, unlike typical 4D reconstruction techniques~\cite{wu20244d,yang2024deformable,li2024spacetime,pumarola2021d,attal2023hyperreel} that can utilize multi-view videos or monocular videos with varying viewpoints, our goal relies on monocular videos typically captured from a fixed viewpoint, which poses a greater challenge on accurate motion and geometry estimation, as demonstrated in Fig.~\ref{fig:ar_recons_ablation}(a). 
To address this, motivated by the fact that videos happen naturally in an autoregressive manner, an object's current state in 3D space can be assumed to be transformed from its prior state. To this end, as shown in Fig.~\ref{fig:framework}(b), we propose to generate current frame's 3D representation based on its previous frame's 3D representation, where the dynamics between adjacent frames are represented by an frame-wise local deformation field, rather than a global deformation field for the whole sequence like previous works~\cite{jiang2023consistent4d, zeng2025stag4d, ren2023dreamgaussian4d}. Such an autoregressive generation manner facilitates more accurate motion modeling by focusing on localized changes, which is able to better capture subtle, frame-to-frame variations, making the generation process more robust and precise. Moreover, as each timestamp provides only a single fixed-viewpoint frame for supervision, the estimated 3D representation may gradually overfit to this frame over the course of training. To mitigate this issue, we introduce a progressive view sampling strategy that utilizes priors from pre-trained large-scale 3D reconstruction models (\eg, LGM~\cite{tang2024lgm}) to progressively provide pseudo views as additional supervisions, which we find can guarantee the spatial-temporal consistency of the underlying geometry to a large extent.

After obtaining each frame's 3D representation, it is observed that due to accumulated errors introduced by autoregressive generation, the 3D representations of later frames exhibit noticeable appearance drift, which affects the quality of the generated results, as demonstrated in Fig.~\ref{fig:refinement_ablation}(a). To address this issue, as shown in Fig.~\ref{fig:framework}(c), we further propose a \textbf{\textit{Refinement}} stage, based on the observation that the geometric structure of each frame remains relatively stable~\cite{niemeyer2022regnerf}. Therefore, we take the 3D representation of the first frame as the canonical space and construct a global deformation field. This field is constrained by the geometric structures of different frames, ensuring that the deformations of the canonical space are kept in check. By doing so, we can significantly reduce appearance drift and guarantee spatial-temporal consistency in the generated 4D assets.

Our main contributions can be summarized as follows:
\begin{itemize}
    \item We propose AR4D, a novel paradigm for generating high-quality 4D assets from monocular videos, bypassing the limitations of Score Distillation Sampling (SDS).

    \item We propose to generate each frame's 3D representation autoregressively using a local deformation field. This process is further improved through a progressive view sampling strategy, enabling precise geometry and motion estimation.

    \item To mitigate the issue of accumulated errors, we propose a refinement stage based on a global deformation field and the extracted geometry of each frame's 3D representation, ensuring the spatial-temporal consistency of generated 4D contents.

    \item Extensive experiments have demonstrated that our proposed AR4D can achieve state-of-the-art performance without SDS, with greater diversity, improved spatial-temporal consistency, and better alignment with input prompts.
    
\end{itemize}

%% file: sec/2_related.tex
\section{Related works: 4D generation}
Building on recent advancements in 3D generation~\cite{poole2022dreamfusion, tang2023dreamgaussian,liu2023syncdreamer,long2024wonder3d} and video generation~\cite{wu2023tune,blattmann2023stable,villegas2022phenaki}, 4D generation has gained substantial interest. Current approaches for 4D generation can be broadly categorized into two types, \ie, generalizable generation methods that leverage prior knowledge, and scene-specific generation methods that optimize for individual scenarios, which we provide a brief introduction here.

\vspace{-4mm}
\paragraph{Generalizable generation methods.}
Prior-based approaches enable 4D generation by either training a generalized model through large-scale multi-modal datasets~\cite{deitke2023objaverse} or integrating pre-trained models directly. For example, methods such as~\cite{xie2024sv4d,li2024vivid,liang2024diffusion4d,zhang20244diffusion} proposed to generate multi-view videos by training a multi-view video diffusion model, which are subsequently processed with 4D reconstruction techniques to produce corresponding 4D assets. To expedite the generation process, L4GM~\cite{ren2024l4gm} introduced the first 4D Large Reconstruction Model capable of producing animated objects in a single feed-forward pass within just one second. Recently, inspired by the powers of video generative models, several approaches~\cite{he2024cameractrl,bahmani2024vd3d,yu2024viewcrafter,xu2024camco,hou2024training} have endowed them with camera control capabilities, allowing for generating videos with varying viewpoints. While photorealistic 4D contents can be achieved, these methods often incur high pre-training costs, and the pre-trained scenes may not be well-suited to the target scene.

\vspace{-4mm}
\paragraph{Scene-specific generation methods.}
Another category of 4D generation methods adopted a scene-specific optimization approach to produce better 4D contents tailored to each individual scene. 
To achieve this, mainstream methods~\cite{jiang2023consistent4d,ling2024align,bahmani20244d,ren2023dreamgaussian4d,zeng2025stag4d,bahmani2025tc4d,gao2024gaussianflow,miao2024pla4d,jiang2024animate3d,yuan20244dynamic,li2024dreammesh4d,zhao2023animate124,zhu2024compositional} primarily distilled knowledge from pre-trained multimodal models (\ie, SDS) to guide the generation process. For instance, Consistent4D~\cite{jiang2023consistent4d} achieved Video-to-4D generation by combining SDS with dynamic NeRF~\cite{mildenhall2021nerf}, followed by a video enhancer to produce high-quality 4D objects. Addressing NeRF's limitations, DreamGaussian4D~\cite{ren2023dreamgaussian4d} introduced the 3DGS~\cite{kerbl20233d} representation, enhanced with texture refinement for fast 4D generation. Recently, STAG4D~\cite{zeng2025stag4d} proposed an innovative approach that can generate anchor multi-view sequences, followed by 4D Gaussian field fitting using SDS to improve 4D generation quality. While these SDS-based methods can achieve reasonable results, they are often hindered by issues~\cite{wang2024prolificdreamer,liang2024luciddreamer,yi2023gaussiandreamer} such as limited diversity, spatial-temporal inconsistency, and poor alignment with input prompts, significantly limiting their practical applications. In contrast, in this paper we propose AR4D, a novel paradigm that is SDS-free for better 4D generation.

%% file: sec/3_preliminaries.tex
\section{Preliminaries: 3DGS}
3D Gaussian Splatting (3DGS)~\cite{kerbl20233d} has shown impressive capability in novel view synthesis, enabling photorealistic novel views to be rendered in real-time. Different from NeRF~\cite{mildenhall2021nerf} that encodes scene properties into neural networks, 3DGS (denoted by $\mathbf{G}$) leverages millions of anisotropic ellipsoids to capture scene geometry and appearance, with each ellipsoid (\ie, 3D Gaussian) parameterized by position $\mathbf{\mu} \in \mathbb{R}^3$, opacity $\alpha \in \mathbb{R}$, covariance $\mathbf{\Sigma} \in \mathbb{R}^{3 \times 3}$ (calculated from scale $\mathbf{s} \in \mathbb{R}^3$ and rotation $\mathbf{r} \in \mathbb{R}^3$), and color $\mathbf{c} \in \mathbb{R}^3$. For simplicity, in this paper we represent the attributes of all ellipsoids collectively as $\mathbf{G} = \{\mathbf{\mu}, \alpha, \mathbf{s}, \mathbf{r}, \mathbf{c}\}$.

To render a novel view, 3DGS employs tile-based rasterization, where 3D Gaussians are projected onto the camera plane to form 2D Gaussians, followed by a point-based rendering operation, which is expressed as:
\begin{equation}
\begin{split}
    &\hat{G}(\mathbf{p},\hat{\mathbf{\mu}}, \hat{\mathbf{\Sigma}}) = e^{-\frac{1}{2}((\mathbf{p}-\hat{\mathbf{\mu}})^T\hat{\mathbf{\Sigma}}^{-1}(\mathbf{p}-\hat{\mathbf{\mu}}))},\\
    \mathbf{c(p)}& = \sum \hat{\mathbf{c}}\hat{\sigma}\prod(1-\hat{\sigma}), \hat{\sigma} = \hat{\alpha}\hat{G}(\mathbf{p},\hat{\mathbf{\mu}}, \hat{\mathbf{\Sigma}}),
\end{split}
\end{equation}
where $\mathbf{c}(\mathbf{p})$ represents the rendered colors of target pixel $\mathbf{p}$, $\hat{G}$ represents projected 2D gaussians whose position, covariance and opacity are $\hat{\mathbf{\mu}}$, $\hat{\mathbf{\Sigma}}$ and $\hat{\alpha}$ respectively.

%% file: sec/4_method.tex
\section{Methods}\label{sec:methods}
For a monocular video $V = \{v_1, v_2, \dots, v_F\}$ (either generated or captured from a fixed viewpoint) with $F$ frames, our objective is to generate its corresponding 4D content without relying on SDS, while enhancing diversity, spatial-temporal consistency, and alignment with the input prompts. As illustrated in Fig.~\ref{fig:framework}, our proposed method follows a three-stage process: \textbf{\textit{Initialization}}, \textbf{\textit{Generation}}, and \textbf{\textit{Refinement}}, each of which is detailed below.

\subsection{Initialization}
\begin{figure}[t]
    \centering
    \begin{subfigure}[b]{0.15\textwidth}
        \centering
        \includegraphics[width=\textwidth]{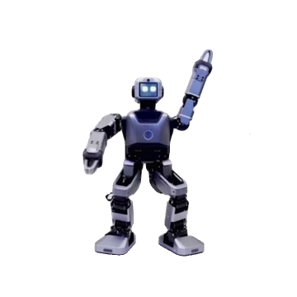}
        \caption{Reference frame.}
    \end{subfigure}
    \hfill
    \begin{subfigure}[b]{0.15\textwidth}
        \centering
        \includegraphics[width=\textwidth]{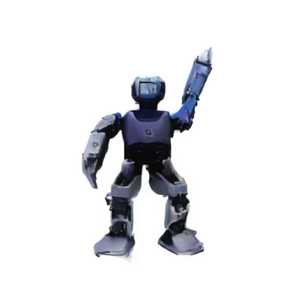}
        \caption{W/o finetuning.}
    \end{subfigure}
    \hfill
    \begin{subfigure}[b]{0.15\textwidth}
        \centering
        \includegraphics[width=\textwidth]{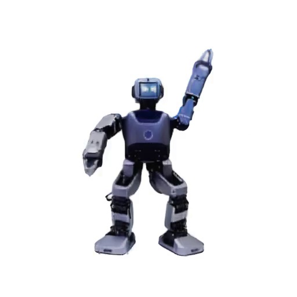}
        \caption{W/. finetuning.}
    \end{subfigure}
    \caption{Ablation studies on finetuning the 3D Gaussians in the \textbf{\textit{Initialization}} stage reveal that finetuning can capture finer texture details in the reference frame, enhancing the quality of subsequent generation.}\label{finetuning in Generation}
\end{figure}

In the first stage, we aim to obtain a 3D representation, which is used to serve as the canonical space for its 4D counterpart. Leveraging recent advances in 3D generation, we first employ a pre-trained multi-view diffusion model to generate several novel views of the first frame, followed by a pre-trained large-scale 3D reconstruction model to recover the corresponding 3D representation (\ie, 3D Gaussians $\mathbf{G}_1^{init} = \{\mathbf{\mu}_1^{init},\alpha_1^{init}, \mathbf{s}_1^{init}, \mathbf{r}_1^{init}, \mathbf{c}_1^{init}\}$) from these generated views. However, as shown in Fig.~\ref{finetuning in Generation}(b), due to the inherent limitations of these pre-trained models, the generated 3D Gaussians often fail to accurately capture the fine-grained texture details of the reference frame $v_1$, presenting additional challenges for the subsequent reconstruction stage. 

To mitigate this issue, we propose a simple yet effective method to fine-tune the obtained 3D Gaussians. Specifically, we keep the parameters $\{\alpha_1^{init}, \mathbf{s}_1^{init}, \mathbf{r}_1^{init}\}$ that influence each gaussian's geometry unchanged, while only optimizing $\{\mathbf{\mu}_1^{init}, \mathbf{c}_1^{init}\}$ to ensure consistency in rendering with the reference frame $v_1$ without harming the overall geometry, using the following equation:
\begin{equation}
    \mathbf{\mu}_1^{ft}, \mathbf{c}_1^{ft} = \underset{\mathbf{\mu}_1^{init}, \mathbf{c}_1^{init}}{\operatorname*{argmin}} \|R^{{ref}}(\mathbf{G}_1^{init}) - v_1\|_2,
\end{equation}
where $R^{{ref}}$ means rendering $\mathbf{G}_1^{init}$ at the view of the reference frame, and the fine-tuned $\mathbf{G}_1$ is thus formulated as $\mathbf{G}_1 =  \{\mathbf{\mu}_1^{ft},\alpha_1^{init}, \mathbf{s}_1^{init}, \mathbf{r}_1^{init}, \mathbf{c}_1^{ft}\}$. 

As shown in Fig.~\ref{finetuning in Generation}(c), the fine-tuned 3D Gaussians can produce results that are better aligned with the reference frame, thereby facilitating the subsequent generation process.

\subsection{Generation}
\begin{figure}[t]
    \centering
    \begin{subfigure}[b]{0.45\textwidth}
        \centering
        \includegraphics[width=\textwidth]{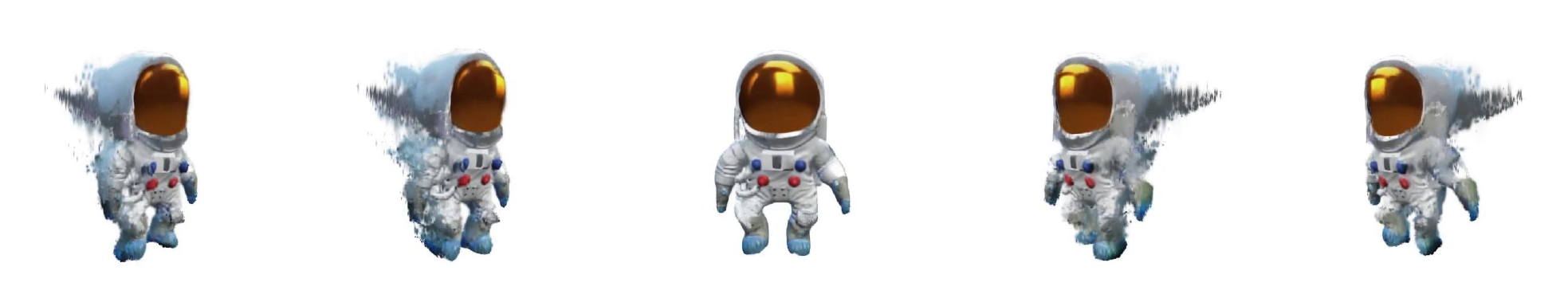}
        \caption{Results when directly applying 4D reconstruction methods.}
    \end{subfigure}
    \hfill
    \begin{subfigure}[b]{0.45\textwidth}
        \centering
        \includegraphics[width=\textwidth]{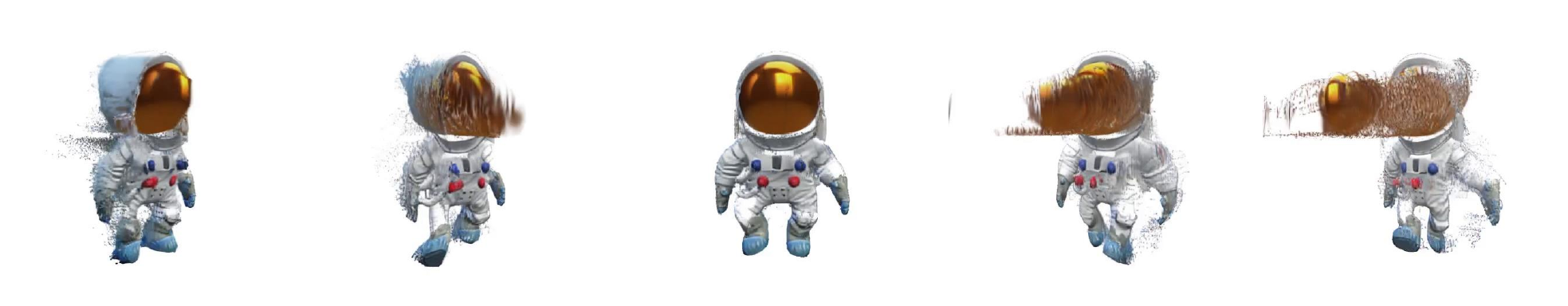}
        \caption{Results when only applying autoregressive 4D generation.}
    \end{subfigure}
    \hfill
    \begin{subfigure}[b]{0.45\textwidth}
        \centering
        \includegraphics[width=\textwidth]{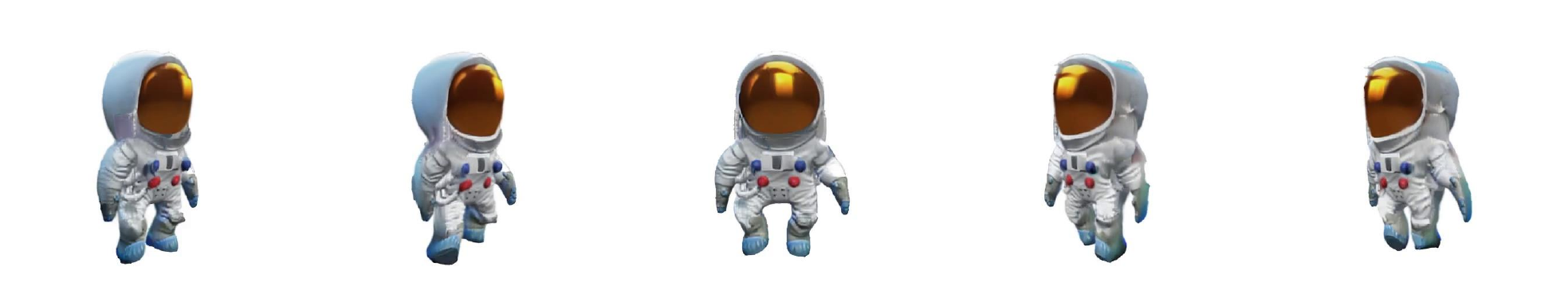}
        \caption{Results when applying autoregressive 4D generation with progressive view sampling.}
    \end{subfigure}
    \caption{Ablation studies on whether applying autoregressive 4D generation and progressive view sampling strategy in the \textbf{\textit{Generation}} stage. With both of them, we can achieve the best performance.} \label{fig:ar_recons_ablation}
\end{figure}

\paragraph{Autoregressive generation.}
To generate the 3D Gaussians for each frame based on $V$ and $\mathbf{G}_1$, a straightforward way is to directly apply common 4D reconstruction methods (\eg, Deform 3DGS~\cite{yang2024deformable}), where $\mathbf{G}_1$ serves as the canonical space, and a global deformation field $F_\theta$ is used to estimate the motion of $\mathbf{G}_1$ at different timestamps by minimizing the difference between the rendered videos and $V$. However, unlike typical 4D reconstruction tasks that can leverage multi-view videos or monocular videos with varying viewpoints, we only have access to monocular videos with a fixed viewpoint, which creates additional challenges for accurate geometry and motion estimation, often resulting in severe artifacts, as demonstrated in Fig.~\ref{fig:ar_recons_ablation}(a).

To address this problem, we propose to leverage the autoregressive nature of videos, which indicates that the 3D Gaussians of consecutive frames undergo only minor deformations. As a result, the 3D Gaussians of the current frame can be seen as being heavily influenced by those of its previous frame. Based on this motivation, we propose to perform the 4D generation from $V$ and $\mathbf{G}_1$ in an autoregressive manner.

Specifically, as shown in Fig.~\ref{fig:framework}(b), for each pair of adjacent frames $v_i$ and $v_{i+1}$, we utilize an independent MLP-based local deformation field $F_{{\theta}_i}$ to model the deformations between their corresponding 3D Gaussians $\mathbf{G}_i = \{\mathbf{\mu}_i,\alpha_i, \mathbf{s}_i, \mathbf{r}_i, \mathbf{c}_i\}$ and $\mathbf{G}_{i+1} = \{\mathbf{\mu}_{i+1},\alpha_{i+1}, \mathbf{s}_{i+1}, \mathbf{r}_{i+1}, \mathbf{c}_{i+1}\}$, which is formulated as follows:
\begin{equation}
\begin{split}
    \{\delta_{\mathbf{\mu}_i}, \delta_{\alpha_i}, \delta_{\mathbf{s}_i}\} = F_{{\theta}_i}(\gamma(\mu_i)), \; 
    \left\{
    \begin{aligned}
        \mathbf{\mu}_{i+1} &= \mathbf{\mu}_{i} + \delta_{\mathbf{\mu}_i} \\
        \alpha_{i+1} &= \alpha_{i} + \delta_{\alpha_i} \\
        \mathbf{s}_{i+1} &= \mathbf{s}_{i} + \delta_{\mathbf{s}_i} \\
        \mathbf{r}_{i+1} &= \mathbf{r}_{i} \\
        \mathbf{c}_{i+1} &= \mathbf{c}_{i} \\
    \end{aligned}
\right.,
\end{split}
\end{equation}
where $\gamma$ is the positional encoding operation that is denoted as follows:
\begin{equation}
\centering
  \gamma(\mathbf{\textit{\textbf{x}}})=(\sin(2^0\mathbf{\textit{\textbf{x}}}),\cos(2^0\mathbf{\textit{\textbf{x}}}),\cdots, \sin(2^{L-1}\mathbf{\textit{\textbf{x}}}),\cos(2^{L-1}\mathbf{\textit{\textbf{x}}})),
\end{equation}
where $L$ is a hyperparameter that is usually set to $10$.

To obtain $\mathbf{G}_{i+1}$ based on $\mathbf{G}_i$, we minimize the difference between the rendered frame $\hat{v}_{i+1} = R^{\text{ref}}(\mathbf{G}_{i+1})$ and the reference frame $v_{i+1}$, as expressed by:
\begin{equation}
\begin{split}
    &\{\theta_i, \mathbf{\mu}_i,\alpha_i, \mathbf{s}_i, \mathbf{r}_i, \mathbf{c}_i\} = \underset{\{\theta_i, \mathbf{\mu}_i,\alpha_i, \mathbf{s}_i, \mathbf{r}_i, \mathbf{c}_i\}}{\operatorname*{argmin}} l_{ref}, \\
    l_{ref} &= \lambda\|\hat{v}_{i+1} - v_{i+1}\|_1 + (1 - \lambda)\text{SSIM}(\hat{v}_{i+1}, v_{i+1})
\end{split}
\end{equation}
where $R^{\text{ref}}$ means rendering $\mathbf{G}_{i+1}$ at the view of $v_{i+1}$, $\text{SSIM}$ means the loss function used to measure the SSIM metric between $\hat{v}_{i+1}$ and $v_{i+1}$, $\lambda$ is a balancing parameter which is set to $0.8$.

\paragraph{Progressive view sampling strategy.} 
As demonstrated in Fig.~\ref{fig:ar_recons_ablation}(b), during the process of autoregressive generation, since each timestamp provides only a single fixed-viewpoint frame for supervision, the generated 3D Gaussians tend to overfit to the reference frames, particularly for the later frames in $V$, leading to significant artifacts in novel views.

To solve this problem, we propose to leverage the powers of pre-trained large-scale 3D reconstruction models~\cite{tang2024lgm} by introducing pseudo novel views as additional supervisions. To achieve this, the major challenge lies on how to obtain appropriate novel views that not only prevent overfitting but also reliable enough to ensure accurate and spatial-temporal-consistent generation.

To this end, we propose a simple yet effective progressive view sampling strategy. Specifically, during the generation process of $\mathbf{G}_{i+1}$, we first render several orthogonal views (including the reference view) of $\mathbf{G}_{i+1}$, which are then fed into the large-scale 3D reconstruction model to create a pseudo 3D Gaussians $\hat{\mathbf{G}}_{i+1}$. Subsequently, considering that during the early stages of optimizing, views rendered by $\hat{\mathbf{G}}_{i+1}$, especially those close to the reference view, are highly reliable, we initially constrain $\mathbf{G}_{i+1}$ by randomly sampling novel views within this close view range using $\hat{\mathbf{G}}_{i+1}$ as additional supervision. With training in progress, the range of sampled viewpoints is progressively expanded to prevent overfitting. 
 
As a result, the progressive view sampling strategy is denoted as follows:
\begin{equation}
    N_{u} = \min(N_{\max}, \lfloor u / \eta \rfloor + N_{start}),
\end{equation}
where $N_{u}$ represents the maximum azimuth angle that can be sampled at the $u$-th iteration, $N_{\max}$ is the upper limit of $N_{u}$, $N_{start}$ is the initial azimuth sampling limit when reconstructing $\mathbf{G}_{i+1}$, and $\eta$ is a hyperparameter controlling the rate at which $N{u}$ increases. During the sampling process, the elevation angle and radius are kept the same as the reference view.
 
Based on this strategy, for a sampled novel view $N_{samp} \sim \mathcal{U}(-N_{u}, N_{u})$, $\mathbf{G}_{i+1}$ is further regularized with the following equations:
\begin{equation}
    \{\theta_i, \mathbf{\mu}_i,\alpha_i, \mathbf{s}_i, \mathbf{r}_i, \mathbf{c}_i\} = \underset{\{\theta_i, \mathbf{\mu}_i,\alpha_i, \mathbf{s}_i, \mathbf{r}_i, \mathbf{c}_i\}}{\operatorname*{argmin}} l_{rgb} + l_{depth},
\end{equation}
where
\begin{equation}
\begin{split}
    l_{rgb} &= \|R^{N_{samp}}(\mathbf{G}_{i+1}) - R^{N_{samp}}(\hat{\mathbf{G}}_{i+1})\|_1 \\
    l_{depth} &= \|R^{N_{samp}}_{depth}(\mathbf{G}_{i+1}) - R^{N_{samp}}_{depth}(\hat{\mathbf{G}}_{i+1})\|_1,
\end{split},
\end{equation}
with $R^{N_{samp}}$ denoting the rendering of images of $\mathbf{G}_{i+1}$ and $\hat{\mathbf{G}}_{i+1}$ at view $N_{samp}$, and $R^{N_{samp}}_{depth}$ representing the rendering of their corresponding depth maps at view $N_{samp}$.

As demonstrated in Fig.~\ref{fig:ar_recons_ablation}(c), the proposed autoregressive generation combined with the progressive view sampling strategy enables accurate motion and geometry estimation significantly.

\subsection{Refinement}\label{sec: refinement}
\begin{figure}[t]
    \centering
    \begin{subfigure}[b]{0.45\textwidth}
        \centering
        \includegraphics[width=\textwidth]{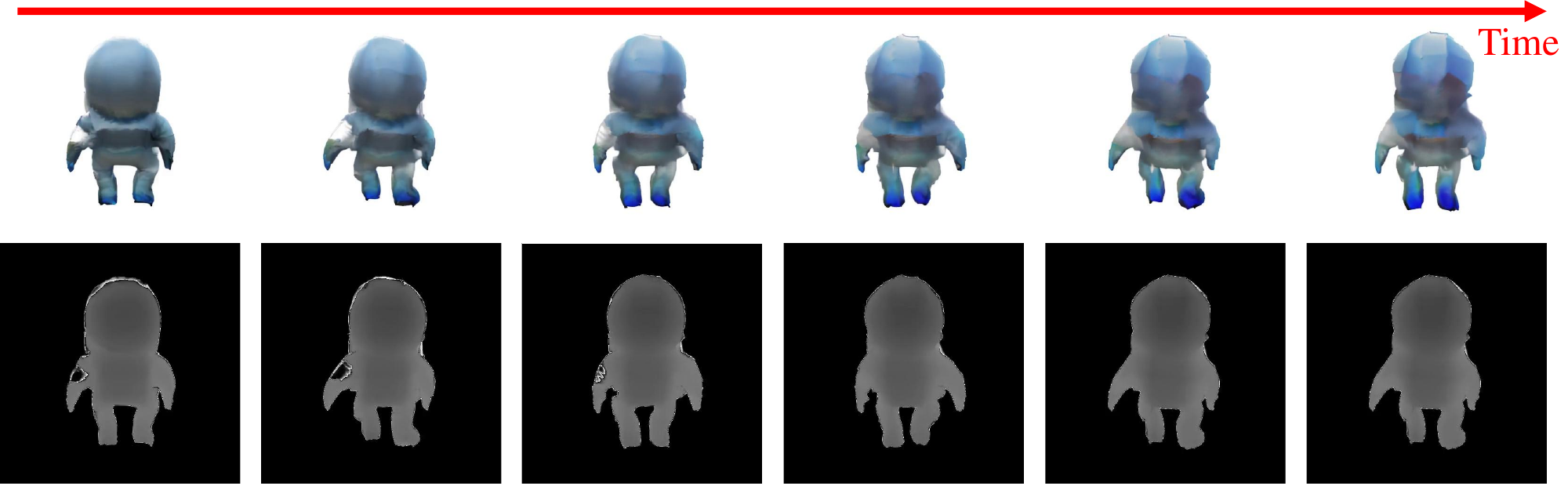}
        \caption{Appearance drift caused by autoregressive generation.}
    \end{subfigure}
    \hfill
    \begin{subfigure}[b]{0.45\textwidth}
        \centering
        \includegraphics[width=\textwidth]{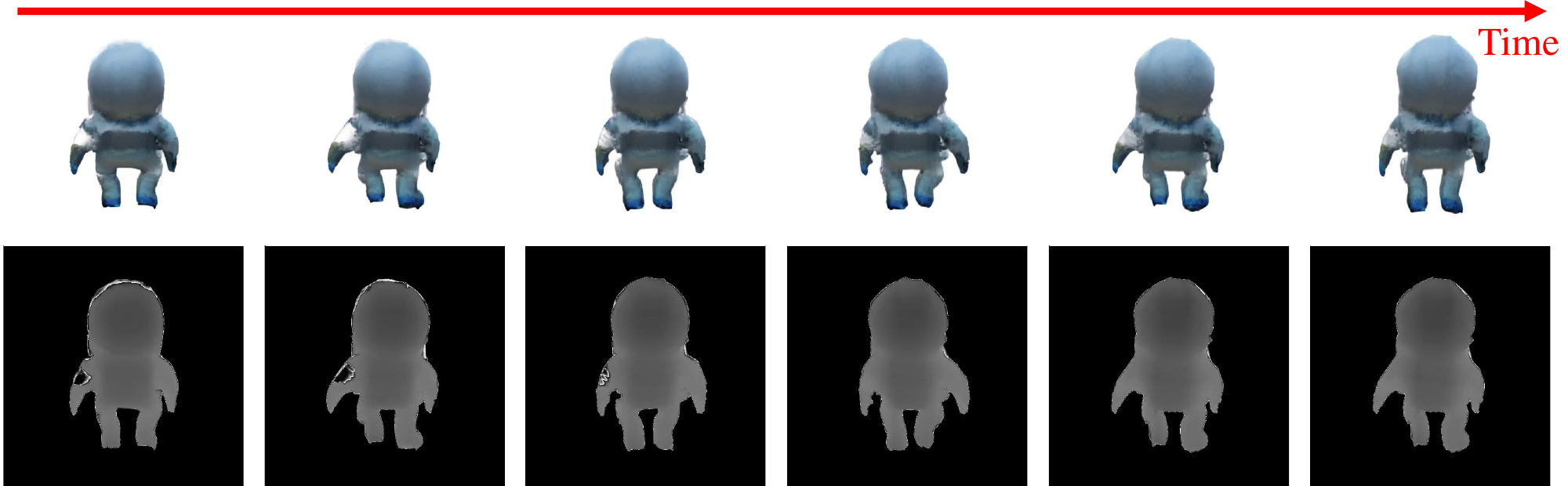}
        \caption{With the refinement stage, no obvious appearance drift is observed.}
    \end{subfigure}
    \caption{Results of the \textbf{\textit{Refinement}} stage demonstrate its effectiveness in addressing appearance drift. While appearance may fluctuate, the geometry (evident in the consistent depth map) remains stable, enabling the generation of spatial-temporal consistent 4D contents.} \label{fig:refinement_ablation}
\end{figure}
As shown in Fig.~\ref{fig:refinement_ablation}(a), performing 4D generation in an autoregressive manner introduces accumulated errors, resulting in noticeable appearance drift, particularly in the later frames of the monocular video $V$.

To address this issue, we propose a refinement stage motivated by the observation that while high-frequency appearance may drift, the geometry (\eg, depth map) of each frame remains relatively low-frequency~\cite{niemeyer2022regnerf} and stable throughout training, as demonstrated in Fig.~\ref{fig:refinement_ablation}. As a result, in this stage, $\mathbf{G}_1 = \{\mathbf{\mu}_1,\alpha_1, \mathbf{s}_1, \mathbf{r}_1, \mathbf{c}_1\}$ is treated as the canonical space, and a global deformation field $F_\theta$, constrained by each frame's depth map, is used to model the deformations across frames, resulting in $\{\mathbf{G}_k^{re} = \{\mathbf{\mu}_k^{re},\alpha_k^{re}, \mathbf{s}_k^{re}, \mathbf{r}_k^{re}, \mathbf{c}_k^{re}\}\}_{k=2}^F$.

Specifically, the relationship between $\mathbf{G}_1$ and $\mathbf{G}_k^{re}$ is formulated as follows:
\begin{equation}
\begin{split}
    \{\delta_{\mathbf{\mu}_k}^{re}, \delta_{\alpha_k}^{re}, \delta_{\mathbf{s}_k}^{re}\} = F_{{\theta}}(\gamma(\mu_1),k), \; 
    \left\{
    \begin{aligned}
        \mathbf{\mu}_{k}^{re} &= \mathbf{\mu}_{1} + \delta_{\mathbf{\mu}_k}^{re} \\
        \alpha_{k}^{re} &= \alpha_{1} + \delta_{\alpha_k}^{re} \\
        \mathbf{s}_{k}^{re} &= \mathbf{s}_{1} + \delta_{\mathbf{s}_k}^{re} \\
        \mathbf{r}_{k}^{re} &= \mathbf{r}_{1} \\
        \mathbf{c}_{k}^{re} &= \mathbf{c}_{1} \\
    \end{aligned}
\right.,
\end{split}
\end{equation}
with $\mathbf{G}_1$ and $F_\theta$ optimized using the following equation:
\begin{equation}
\begin{split}
    &\{\theta, \mathbf{\mu}_1,\alpha_1, \mathbf{s}_1, \mathbf{r}_1, \mathbf{c}_1\} = \underset{\{\theta, \mathbf{\mu}_1,\alpha_1, \mathbf{s}_1, \mathbf{r}_1, \mathbf{c}_1\}}{\operatorname*{argmin}} l_{ref}^{re} + l_{depth}^{re}
\end{split},
\end{equation}
where 
\begin{equation}
\begin{split}
    l_{ref}^{re} &= \mathbb{E}_{k}[\|R^{ref}(\mathbf{G}_{k}) - R^{ref}({\mathbf{G}}_{k}^{re})\|_1] \\
    l_{depth}^{re} &= \mathbb{E}_{k}[\|R^{N_{samp}^{re}}_{depth}(\mathbf{G}_{k}) - R^{N_{samp}^{re}}_{depth}({\mathbf{G}}_{k}^{re})\|_1],
\end{split},
\end{equation}
$\mathbf{G}_k$ denotes the 3D Gaussians obtained during the reconstruction stage for the $k$-th frame, $R^{ref}$ denotes rendering of $\mathbf{G}_k$ and $\mathbf{G}_k^{re}$ at view of the reference frame $v_k$, $N_{samp}^{re}$ refers to a randomly sampled viewpoint within the view space, and $R^{N_{samp}^{re}}_{depth}$ represents the rendering of the depth maps of $\mathbf{G}_k$ and $\mathbf{G}_k^{re}$ from the viewpoint $N_{samp}^{re}$.

As demonstrated in Fig.~\ref{fig:refinement_ablation}(b), this refinement ensures that each frame's geometry, obtained in the \textbf{\textit{Generation}} stage, remains unchanged while its appearance is directly deformed from the same 3D Gaussians $\mathbf{G}_1$, preventing significant appearance drift and thus improving spatial-temporal consistency.

%% file: sec/5_experiments.tex
\section{Experiments}
\subsection{Experimental settings}
\paragraph{Implementation details.}
\begin{figure}[t]
    \centering
    \includegraphics[width=1\linewidth]{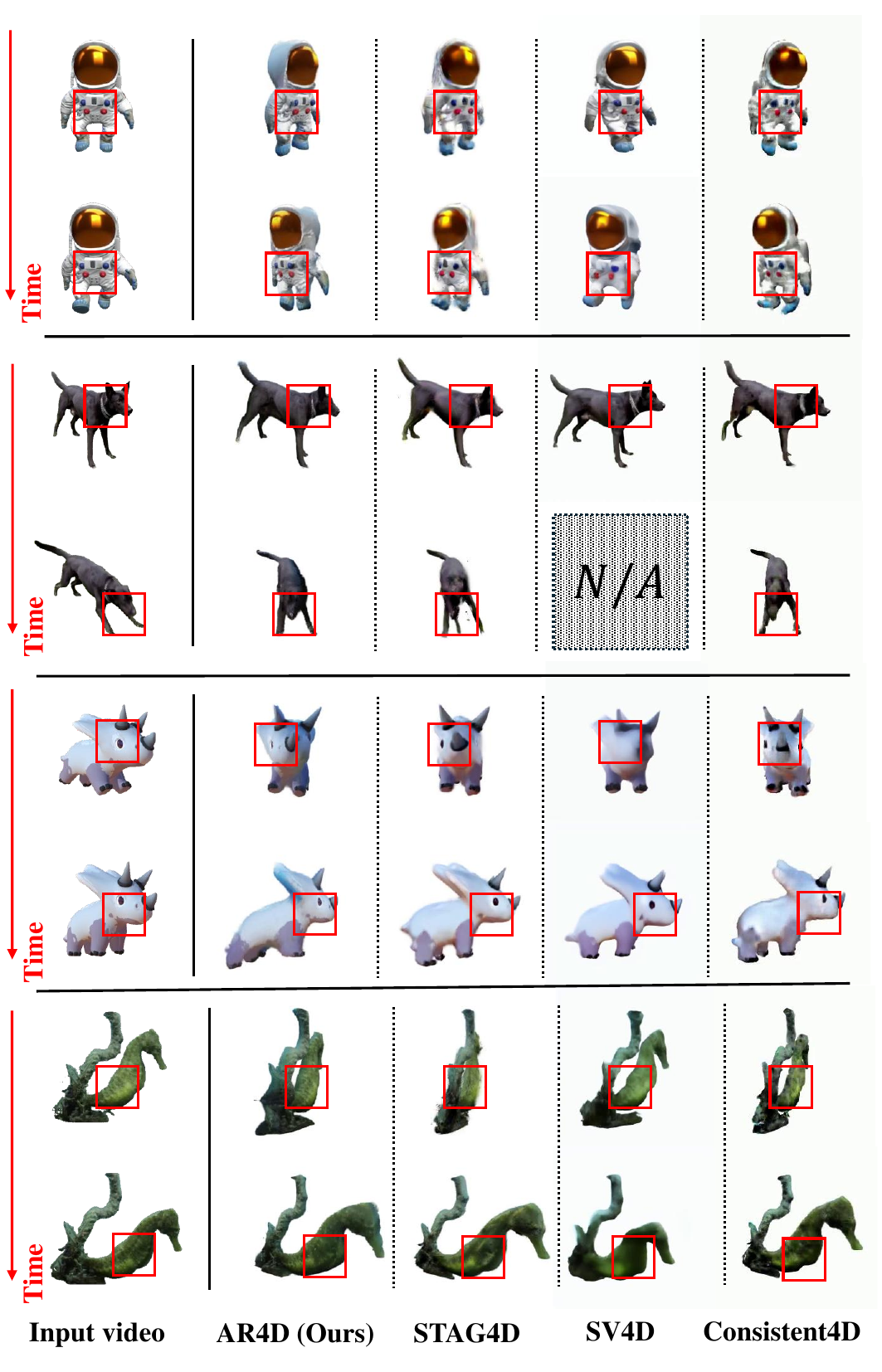}
    \caption{Comparisons of our proposed AR4D with other state-of-the-art methods on Video-to-4D. Our proposed method can generate more detailed results with improved alignment to input prompts. $\textbf{\textit{N/A}}$ indicates that the corresponding method fails to generate novel views for the current frame.}
\label{fig:main_results_video24d}
\end{figure}
\begin{table}
    \resizebox{\linewidth}{!}{\input{tables/results_video24d}}
    \caption{Quantitative comparisons of our method with other state-of-the-art methods on Video-to-4D.
    The best, second-best, and third-best entries are marked in red, orange, and yellow, respectively.}
    \label{tab:Quantitative_video24d}
\end{table}

\begin{figure}[t]
    \centering
    \includegraphics[width=1\linewidth]{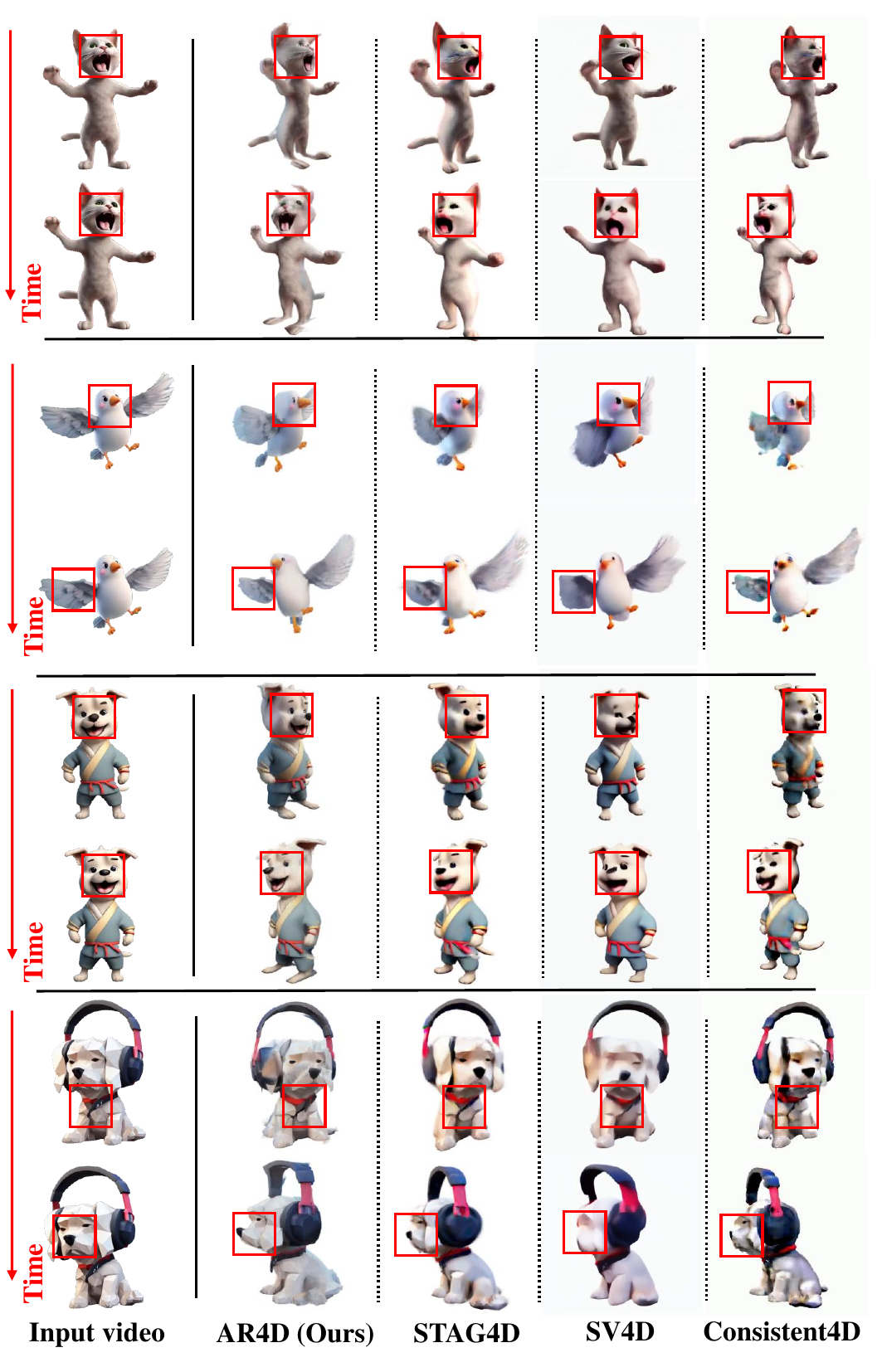}
    \caption{{Comparisons of our proposed AR4D with other state-of-the-art methods on Text-to-4D. Our proposed method can generate clearer results with enhanced spatial-temporal consistency.}}
\label{fig:main_results_text24d}
\end{figure}
\begin{table}
    \resizebox{\linewidth}{!}{\input{tables/results_text24d}}
    \caption{Quantitative comparisons of our method with other state-of-the-art methods on Text-to-4D.
    }
    \label{tab:Quantitative_text24d}
\end{table}

To generate 4D objects, during the \textbf{\textit{Initialization}} stage, we use MVDream~\cite{shi2023mvdream} as the multi-view diffusion model, followed by LGM~\cite{tang2024lgm} as the large-scale 3D reconstruction model to generate 3D Gaussians for the first frame. In the \textbf{\textit{Generation}} stage, each local deformation MLP, applied to adjacent frame pairs, consists of 8 hidden layers with 256 hidden units per layer. The initial learning rate for these MLPs is set to $5 \times 10^{-4}$ and decays to $1 \times 10^{-6}$ by the end of training. For the progressive view sampling strategy, $N_{max}$ is 180, $N_{start}$ is 1, and $\eta$ is set to 10. The training iterations for each $F_{\theta_i}$ are set to 2000. In the \textbf{\textit{Refinement}} stage, the global deformation field $F_\theta$ also has 8 hidden layers with 256 hidden units per layer, with an initial learning rate of $5 \times 10^{-4}$ decaying to $1 \times 10^{-6}$. The total training iterations for $F_\theta$ is set to 30,000. To demonstrate the superiority of our method, we also perform experiments on the task of 4D scene generation, using Splatt3R~\cite{smart2024splatt3r} as both the single-view generator and 3D reconstruction model. We perform our experiments on a single NVIDIA A100 GPU. Please see more details in the supplementary materials.

\vspace{-4mm}
\paragraph{Datasets and metrics.}
Following the experimental protocols outlined by STAG4D~\cite{zeng2025stag4d}, we use the provided datasets to conduct experiments on both video-to-4D and image-to-4D tasks. The image-to-4D task is performed in two steps: first, converting the image to video, followed by the video-to-4D transformation. To evaluate the quality of the generated results, we report PSNR, SSIM~\cite{wang2004image}, and LPIPS~\cite{zhang2018unreasonable} to assess the alignment between the rendered videos and the ground truth. Additionally, we report CLIP similarity~\cite{radford2021learning} and FVD scores~\cite{unterthiner2018towards} to measure the consistency between the rendered novel views and the reference views. Kindly refer to supplementary materials for more details.

\vspace{-4mm}
\paragraph{Baselines.}
For 4D object generation, we compare our proposed AR4D with several state-of-the-art methods, including Consistent4D~\cite{jiang2023consistent4d}, SV4D~\cite{xie2024sv4d}, and STAG4D~\cite{zeng2025stag4d}. As few methods exist for 4D scene generation from monocular videos with a fixed viewpoint, we compare our method with a commonly used 4D reconstruction approach, \ie, Deform 3DGS~\cite{yang2024deformable}.

\subsection{Comparisons with state-of-the-art methods}
\begin{figure}[t]
    \centering
    \includegraphics[width=1\linewidth]{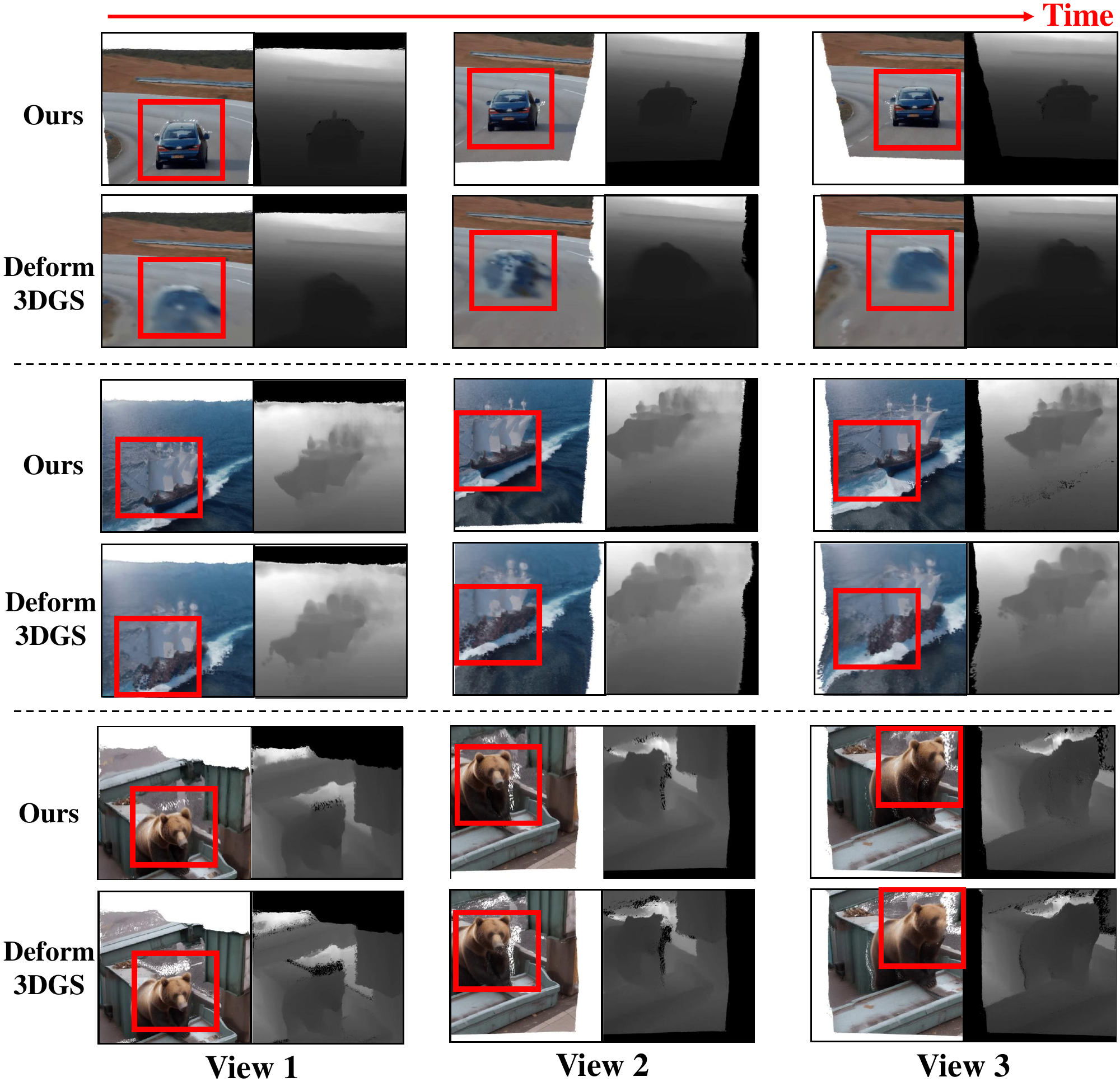}
    \caption{{Comparisons of our proposed AR4D with Deform 3DGS~\cite{yang2024deformable} on 4D scene generation.}}
    \label{fig:GENERATED_SCENES_COMPARISONS}
\end{figure}
\paragraph{Qualitative comparisons.}
As shown in Fig.~\ref{fig:main_results_video24d} and Fig.~\ref{fig:main_results_text24d}, given a monocular video, Consistent4D produces over-saturated outputs with a blurred appearance, limited by the intrinsic constraints of SDS. Similarly, although STAG4D can reduce over-saturation to some degree, the results still exhibit noticeable noise and unrealistic, fabricated patterns.
For SV4D, as a general 4D generative model, the domain gap issue leads to highly blurred novel views, restricting it to processing short input videos of only 21 frames. In contrast, our proposed AR4D can achieve clearer results with enhanced alignment to input videos and improved spatial-temporal consistency. For 4D scene generation, as shown in Fig.~\ref{fig:GENERATED_SCENES_COMPARISONS}, since only monocular videos with a fixed viewpoint are available, Deform 3DGS  struggles to estimate accurate geometry and motion, leading to degradation outputs. Conversely, our method can overcome this problem and achieve more reasonable results. We provide more visualizations in the supplementary materials.

\vspace{-6mm}
\paragraph{Quantitative comparisons.} As demonstrated in Tab.~\ref{tab:Quantitative_video24d} and Tab.~\ref{tab:Quantitative_text24d}, our proposed method can achieve the highest performance, with an average improvement of 1 dB in PSNR, demonstrating that AR4D can generate 4D assets closely aligned with the input. Moreover, we can also achieve the best CLIP similarity and FVD-score, indicating superior spatial-temporal consistency in the generated 4D objects.
\subsection{Ablation studies}\label{sec: ablations}
\begin{figure}[t]
    \centering
    \begin{subfigure}[b]{0.45\textwidth}
        \centering
        \includegraphics[width=\textwidth]{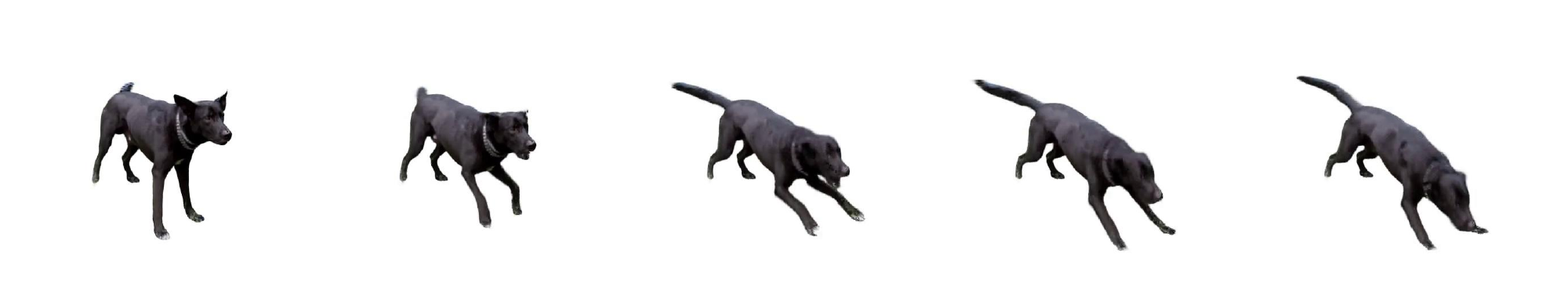}
        \vspace{-8mm}
        \caption{Input monocular video.}
    \end{subfigure}
    \hfill
    \begin{subfigure}[b]{0.45\textwidth}
        \centering
        \includegraphics[width=\textwidth]{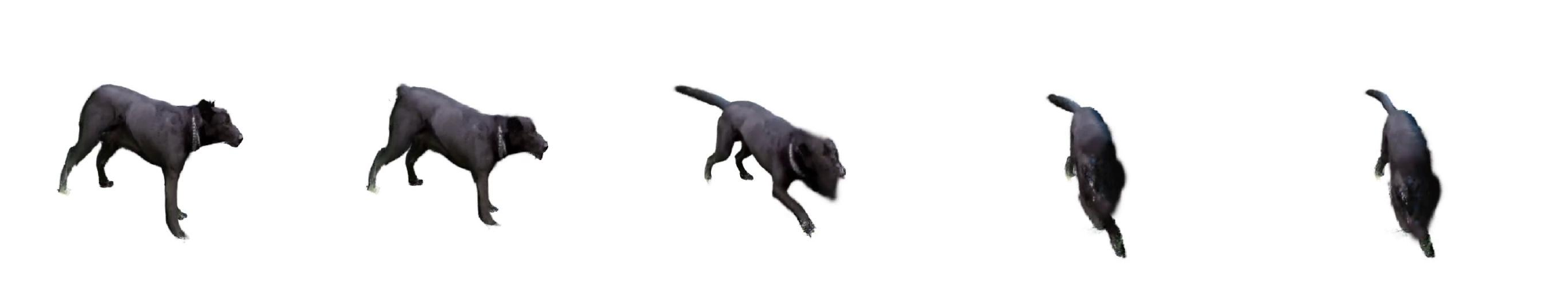}
        \vspace{-8mm}
        \caption{Results rendered without autoregressive generation.}
    \end{subfigure}
    \hfill
    \begin{subfigure}[b]{0.45\textwidth}
        \centering
        \includegraphics[width=\textwidth]{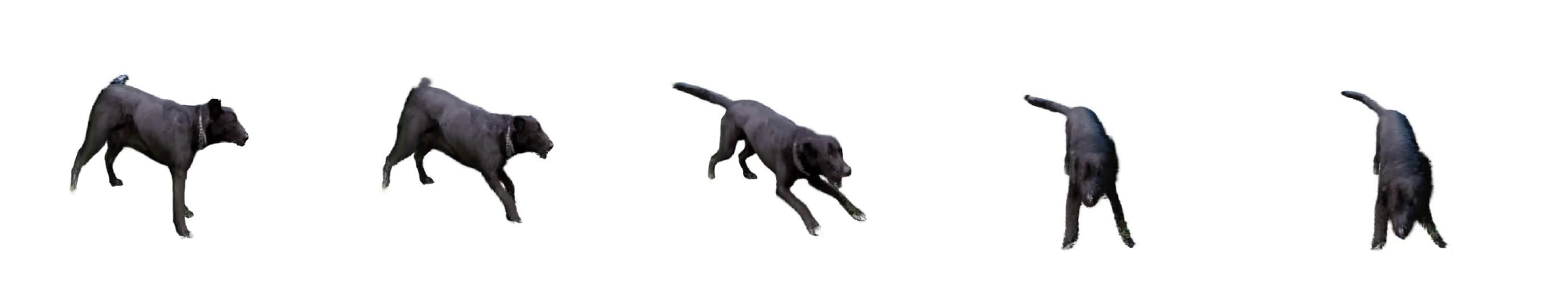}
        \vspace{-8mm}
        \caption{Results rendered with autoregressive generation.}
    \end{subfigure}
    \caption{Ablation study on the effect of autoregressive generation: incorporating it improves motion and geometry estimation, leading to more realistic results.} 
    \label{fig:ablation_no_ar}
\end{figure}

\vspace{-2mm}
\begin{table}
    \resizebox{\linewidth}{!}{\input{tables/ablation_results}}
    \caption{We perform ablation studies on the Video-to-4D dataset, where \textbf{Init-ft} means finetuning the 3D Gaussians obtained in the \textbf{\textit{Initialization}} stage, \textbf{AR} and \textbf{PVS} means autoregressive generation and progressive view sampling strategy in the \textbf{\textit{Generation}} stage, \textbf{Refine} means whether incorporating the \textbf{\textit{Refinement}} stage.
    } 
    \label{tab:Ablation results}
\end{table}
To showcase the effectiveness of our design choices, we
conduct both quantitative and qualitative ablation studies on the task of video-to-4D. As shown in Tab.~\ref{tab:Ablation results} and Fig.~\ref{finetuning in Generation},  when omitting finetuning the 3D Gaussians obtained in the \textbf{\textit{Initialization}} stage, a performance drop is observed due to the inherent limitations of adopted pre-trained 3D generators. Similarly, when removing the refinement stage, both alignment with input videos and spatial-temporal consistency are negatively influenced, owning to the appearance drift mentioned in Sec.~\ref{sec: refinement} and Fig.~\ref{fig:refinement_ablation}. As demonstrated in Fig.~\ref{fig:ar_recons_ablation}, if we remove the progressive view sampling strategy, the generated 4D assets overfit to input videos, resulting in relatively high reconstruction metrics (e.g., PSNR) but significantly lower FVD scores. Additionally, as demonstrated in Fig.~\ref{fig:ablation_no_ar}, if we remove the autoregressive generation, the performance also drops due to the lack of precise motion and geometry estimation. More visualizations are provided in the supplementary materials.

\subsection{Benefits of AR4D for 4D generation}
\vspace{-2mm}
\textbf{1) Greater diversity:} By directly using outputs from any generative model to guide the generation process, our approach can fully inherit the diversity of these pre-trained models, avoiding the low-diversity problem caused by SDS~\cite{wang2024prolificdreamer}; \textbf{2) Enhanced spatial-temporal consistency: }The autoregressive nature of our proposed AR4D facilitates precise estimation of the 4D asset's motion and appearance, resulting in enhanced spatial-temporal consistency; \textbf{3) Better alignment with input prompts: }By directly supervising the generation process in pixel space and bypassing the indirect feature-space supervision of SDS, we can avoid the blurriness introduced by SDS~\cite{liang2024luciddreamer}, leading to outputs that are more closely aligned with the input prompts.

%% file: tables/results_video24d.tex
\begin{tabular}{c|c|c|c|c}
\hline
\hline
{Method} & {Consistent4D}~\cite{jiang2023consistent4d} & {SV4D}~\cite{xie2024sv4d} & {STAG4D}~\cite{zeng2025stag4d} & \textbf{Ours} \\
\hline
PSNR$\uparrow$ & 24.77 & \cellcolor{yellow!25}28.23 & \cellcolor{orange!25}29.91 & \cellcolor{red!25}\textbf{31.00} \\
SSIM$\uparrow$ & 0.91 & \cellcolor{yellow!25}0.92 & \cellcolor{orange!25}0.94 & \cellcolor{red!25}\textbf{0.97} \\
LPIPS$\downarrow$ & 0.11 & \cellcolor{yellow!25}0.06 & \cellcolor{orange!25}0.05 & \cellcolor{red!25}\textbf{0.02} \\
\hline
CLIP-S$\uparrow$ & \cellcolor{orange!25}0.90 & \cellcolor{yellow!25}0.89 & \cellcolor{orange!25}0.90 & \cellcolor{red!25}\textbf{0.92} \\
FVD$\downarrow$ & \cellcolor{yellow!25}873 & - & \cellcolor{orange!25}737 & \cellcolor{red!25}\textbf{617} \\
FVD-16$\downarrow$ & 611 & \cellcolor{yellow!25}592 & \cellcolor{orange!25}573 & \cellcolor{red!25}\textbf{478} \\
\hline
\hline
\end{tabular}

%% file: tables/results_text24d.tex
\begin{tabular}{c|c|c|c|c}
\hline
\hline
{Method} & {Consistent4D}~\cite{jiang2023consistent4d} & {SV4D}~\cite{xie2024sv4d} & {STAG4D}~\cite{zeng2025stag4d} & \textbf{Ours} \\
\hline
PSNR$\uparrow$ & 23.38 & \cellcolor{yellow!25}28.34 & \cellcolor{orange!25}30.26 & \cellcolor{red!25}\textbf{31.06} \\
SSIM$\uparrow$ & 0.88 & \cellcolor{yellow!25}0.93 & \cellcolor{orange!25}0.95 & \cellcolor{red!25}\textbf{0.98} \\
LPIPS$\downarrow$ & 0.17 & \cellcolor{yellow!25}0.09 & \cellcolor{orange!25}0.06 & \cellcolor{red!25}\textbf{0.03} \\
\hline
CLIP-S$\uparrow$ & \cellcolor{yellow!25}0.89 & 0.88 & \cellcolor{orange!25}0.90 & \cellcolor{red!25}\textbf{0.92} \\
FVD$\downarrow$ & \cellcolor{yellow!25}1250 & - & \cellcolor{orange!25}1065 & \cellcolor{red!25}\textbf{890} \\
FVD-16$\downarrow$ & 1084 & \cellcolor{yellow!25}1032 & \cellcolor{orange!25}947 & \cellcolor{red!25}\textbf{681} \\
\hline
\hline
\end{tabular}

%% file: tables/ablation_results.tex
\begin{tabular}{c|c c c c c c}
\hline 
\hline 
\textbf{Init-ft} & \ding{51} & \ding{55} & \ding{51} & \ding{51} & \ding{51} & \ding{51} \\ 
\textbf{AR} & \ding{51} & \ding{51} & \ding{51} & \ding{51} & \ding{55} & \ding{51} \\ 
\textbf{PVS} & \ding{51} & \ding{51} & \ding{51} & \ding{55} & \ding{51} & \ding{55} \\ 
\textbf{Refine} & \ding{51} & \ding{51} & \ding{55} & \ding{51} & \ding{51} & \ding{55} \\ \hline
PSNR$\uparrow$ & \cellcolor{red!25}\textbf{31.00} & 30.43 & 30.24 & \cellcolor{orange!25}30.86 & 30.53 & \cellcolor{yellow!25}30.74\\ 
SSIM$\uparrow$ & \cellcolor{red!25}\textbf{0.97} & \cellcolor{yellow!25}0.95 & \cellcolor{yellow!25}0.95 & \cellcolor{orange!25}0.96 & 0.94 & \cellcolor{yellow!25}0.95 \\ 
LPIPS$\downarrow$ & \cellcolor{red!25}\textbf{0.02} & \cellcolor{orange!25}0.04 & \cellcolor{yellow!25}0.08 & \cellcolor{orange!25}0.04 & 0.10 & 0.10 \\ 
FVD$\downarrow$ & \cellcolor{red!25}\textbf{617} & \cellcolor{orange!25}681 & \cellcolor{yellow!25}712 & 1532 & 1026 & 1637 \\ 
\hline
\hline
\end{tabular}

%% file: sec/6_discussion.tex
\section{Conclusion}
\vspace{-2mm}
In this paper, we introduce AR4D, a novel approach for SDS-free 4D generation from monocular videos. AR4D operates in three stages: \textbf{\textit{1) Initialization:}} Pre-trained 3D generators are employed to extract 3D Gaussians from the video's first frame, which are then fine-tuned to establish the canonical space for its 4D counterpart. \textbf{\textit{2) Generation:}} For more accurate motion and geometry estimation, 3D Gaussians are generated for each frame in an autoregressive manner, complemented by a progressive view sampling strategy to mitigate overfitting. \textbf{\textit{3) Refinement:}} To counteract appearance drift introduced by autoregressive generation, a global deformation field works in conjunction with per-frame geometry to achieve detailed refinement. Experiments have demonstrated that our method can achieve state-of-the-art 4D generation, with greater diversity, improved spatial-temporal consistency, and better alignment with input prompts.

%% file: main.bbl
\begin{thebibliography}{58}
\providecommand{\natexlab}[1]{#1}
\providecommand{\url}[1]{\texttt{#1}}
\expandafter\ifx\csname urlstyle\endcsname\relax
  \providecommand{\doi}[1]{doi: #1}\else
  \providecommand{\doi}{doi: \begingroup \urlstyle{rm}\Url}\fi

\bibitem[Attal et~al.(2023)Attal, Huang, Richardt, Zollhoefer, Kopf, O’Toole, and Kim]{attal2023hyperreel}
Benjamin Attal, Jia-Bin Huang, Christian Richardt, Michael Zollhoefer, Johannes Kopf, Matthew O’Toole, and Changil Kim.
\newblock Hyperreel: High-fidelity 6-dof video with ray-conditioned sampling.
\newblock In \emph{Proceedings of the IEEE/CVF Conference on Computer Vision and Pattern Recognition}, pages 16610--16620, 2023.

\bibitem[Bahmani et~al.(2024{\natexlab{a}})Bahmani, Skorokhodov, Rong, Wetzstein, Guibas, Wonka, Tulyakov, Park, Tagliasacchi, and Lindell]{bahmani20244d}
Sherwin Bahmani, Ivan Skorokhodov, Victor Rong, Gordon Wetzstein, Leonidas Guibas, Peter Wonka, Sergey Tulyakov, Jeong~Joon Park, Andrea Tagliasacchi, and David~B Lindell.
\newblock 4d-fy: Text-to-4d generation using hybrid score distillation sampling.
\newblock In \emph{Proceedings of the IEEE/CVF Conference on Computer Vision and Pattern Recognition}, pages 7996--8006, 2024{\natexlab{a}}.

\bibitem[Bahmani et~al.(2024{\natexlab{b}})Bahmani, Skorokhodov, Siarohin, Menapace, Qian, Vasilkovsky, Lee, Wang, Zou, Tagliasacchi, et~al.]{bahmani2024vd3d}
Sherwin Bahmani, Ivan Skorokhodov, Aliaksandr Siarohin, Willi Menapace, Guocheng Qian, Michael Vasilkovsky, Hsin-Ying Lee, Chaoyang Wang, Jiaxu Zou, Andrea Tagliasacchi, et~al.
\newblock Vd3d: Taming large video diffusion transformers for 3d camera control.
\newblock \emph{arXiv preprint arXiv:2407.12781}, 2024{\natexlab{b}}.

\bibitem[Bahmani et~al.(2025)Bahmani, Liu, Yifan, Skorokhodov, Rong, Liu, Liu, Park, Tulyakov, Wetzstein, et~al.]{bahmani2025tc4d}
Sherwin Bahmani, Xian Liu, Wang Yifan, Ivan Skorokhodov, Victor Rong, Ziwei Liu, Xihui Liu, Jeong~Joon Park, Sergey Tulyakov, Gordon Wetzstein, et~al.
\newblock Tc4d: Trajectory-conditioned text-to-4d generation.
\newblock In \emph{European Conference on Computer Vision}, pages 53--72. Springer, 2025.

\bibitem[Blattmann et~al.(2023)Blattmann, Dockhorn, Kulal, Mendelevitch, Kilian, Lorenz, Levi, English, Voleti, Letts, et~al.]{blattmann2023stable}
Andreas Blattmann, Tim Dockhorn, Sumith Kulal, Daniel Mendelevitch, Maciej Kilian, Dominik Lorenz, Yam Levi, Zion English, Vikram Voleti, Adam Letts, et~al.
\newblock Stable video diffusion: Scaling latent video diffusion models to large datasets.
\newblock \emph{arXiv preprint arXiv:2311.15127}, 2023.

\bibitem[Deitke et~al.(2023)Deitke, Schwenk, Salvador, Weihs, Michel, VanderBilt, Schmidt, Ehsani, Kembhavi, and Farhadi]{deitke2023objaverse}
Matt Deitke, Dustin Schwenk, Jordi Salvador, Luca Weihs, Oscar Michel, Eli VanderBilt, Ludwig Schmidt, Kiana Ehsani, Aniruddha Kembhavi, and Ali Farhadi.
\newblock Objaverse: A universe of annotated 3d objects.
\newblock In \emph{Proceedings of the IEEE/CVF Conference on Computer Vision and Pattern Recognition}, pages 13142--13153, 2023.

\bibitem[Gao et~al.(2024)Gao, Xu, Cao, Mildenhall, Ma, Chen, Tang, and Neumann]{gao2024gaussianflow}
Quankai Gao, Qiangeng Xu, Zhe Cao, Ben Mildenhall, Wenchao Ma, Le Chen, Danhang Tang, and Ulrich Neumann.
\newblock Gaussianflow: Splatting gaussian dynamics for 4d content creation.
\newblock \emph{arXiv preprint arXiv:2403.12365}, 2024.

\bibitem[He et~al.(2024)He, Xu, Guo, Wetzstein, Dai, Li, and Yang]{he2024cameractrl}
Hao He, Yinghao Xu, Yuwei Guo, Gordon Wetzstein, Bo Dai, Hongsheng Li, and Ceyuan Yang.
\newblock Cameractrl: Enabling camera control for text-to-video generation.
\newblock \emph{arXiv preprint arXiv:2404.02101}, 2024.

\bibitem[Hou et~al.(2024)Hou, Wei, Zeng, and Chen]{hou2024training}
Chen Hou, Guoqiang Wei, Yan Zeng, and Zhibo Chen.
\newblock Training-free camera control for video generation.
\newblock \emph{arXiv preprint arXiv:2406.10126}, 2024.

\bibitem[Jiang et~al.(2024{\natexlab{a}})Jiang, Yu, Cao, Wang, Hu, and Gao]{jiang2024animate3d}
Yanqin Jiang, Chaohui Yu, Chenjie Cao, Fan Wang, Weiming Hu, and Jin Gao.
\newblock Animate3d: Animating any 3d model with multi-view video diffusion.
\newblock \emph{arXiv preprint arXiv:2407.11398}, 2024{\natexlab{a}}.

\bibitem[Jiang et~al.(2024{\natexlab{b}})Jiang, Zhang, Gao, Hu, and Yao]{jiang2023consistent4d}
Yanqin Jiang, Li Zhang, Jin Gao, Weimin Hu, and Yao Yao.
\newblock Consistent4d: Consistent 360 $\{$$\backslash$deg$\}$ dynamic object generation from monocular video.
\newblock \emph{The Twelfth International Conference on Learning Representations}, 2024{\natexlab{b}}.

\bibitem[Kerbl et~al.(2023)Kerbl, Kopanas, Leimk{\"u}hler, and Drettakis]{kerbl20233d}
Bernhard Kerbl, Georgios Kopanas, Thomas Leimk{\"u}hler, and George Drettakis.
\newblock 3d gaussian splatting for real-time radiance field rendering.
\newblock \emph{ACM Trans. Graph.}, 42\penalty0 (4):\penalty0 139--1, 2023.

\bibitem[Li et~al.(2024{\natexlab{a}})Li, Zheng, Zhu, Mai, Zhang, Wonka, and Ghanem]{li2024vivid}
Bing Li, Cheng Zheng, Wenxuan Zhu, Jinjie Mai, Biao Zhang, Peter Wonka, and Bernard Ghanem.
\newblock Vivid-zoo: Multi-view video generation with diffusion model.
\newblock \emph{arXiv preprint arXiv:2406.08659}, 2024{\natexlab{a}}.

\bibitem[Li et~al.(2024{\natexlab{b}})Li, Chen, and Liu]{li2024dreammesh4d}
Zhiqi Li, Yiming Chen, and Peidong Liu.
\newblock Dreammesh4d: Video-to-4d generation with sparse-controlled gaussian-mesh hybrid representation.
\newblock \emph{arXiv preprint arXiv:2410.06756}, 2024{\natexlab{b}}.

\bibitem[Li et~al.(2024{\natexlab{c}})Li, Chen, Li, and Xu]{li2024spacetime}
Zhan Li, Zhang Chen, Zhong Li, and Yi Xu.
\newblock Spacetime gaussian feature splatting for real-time dynamic view synthesis.
\newblock In \emph{Proceedings of the IEEE/CVF Conference on Computer Vision and Pattern Recognition}, pages 8508--8520, 2024{\natexlab{c}}.

\bibitem[Liang et~al.(2024{\natexlab{a}})Liang, Yin, Xu, Liang, Wang, Plataniotis, Zhao, and Wei]{liang2024diffusion4d}
Hanwen Liang, Yuyang Yin, Dejia Xu, Hanxue Liang, Zhangyang Wang, Konstantinos~N Plataniotis, Yao Zhao, and Yunchao Wei.
\newblock Diffusion4d: Fast spatial-temporal consistent 4d generation via video diffusion models.
\newblock \emph{arXiv preprint arXiv:2405.16645}, 2024{\natexlab{a}}.

\bibitem[Liang et~al.(2024{\natexlab{b}})Liang, Yang, Lin, Li, Xu, and Chen]{liang2024luciddreamer}
Yixun Liang, Xin Yang, Jiantao Lin, Haodong Li, Xiaogang Xu, and Yingcong Chen.
\newblock Luciddreamer: Towards high-fidelity text-to-3d generation via interval score matching.
\newblock In \emph{Proceedings of the IEEE/CVF Conference on Computer Vision and Pattern Recognition}, pages 6517--6526, 2024{\natexlab{b}}.

\bibitem[Ling et~al.(2024)Ling, Kim, Torralba, Fidler, and Kreis]{ling2024align}
Huan Ling, Seung~Wook Kim, Antonio Torralba, Sanja Fidler, and Karsten Kreis.
\newblock Align your gaussians: Text-to-4d with dynamic 3d gaussians and composed diffusion models.
\newblock In \emph{Proceedings of the IEEE/CVF Conference on Computer Vision and Pattern Recognition}, pages 8576--8588, 2024.

\bibitem[Liu et~al.(2024)Liu, Xu, Jin, Chen, Varma~T, Xu, and Su]{liu2024one}
Minghua Liu, Chao Xu, Haian Jin, Linghao Chen, Mukund Varma~T, Zexiang Xu, and Hao Su.
\newblock One-2-3-45: Any single image to 3d mesh in 45 seconds without per-shape optimization.
\newblock \emph{Advances in Neural Information Processing Systems}, 36, 2024.

\bibitem[Liu et~al.(2023{\natexlab{a}})Liu, Wu, Van~Hoorick, Tokmakov, Zakharov, and Vondrick]{liu2023zero}
Ruoshi Liu, Rundi Wu, Basile Van~Hoorick, Pavel Tokmakov, Sergey Zakharov, and Carl Vondrick.
\newblock Zero-1-to-3: Zero-shot one image to 3d object.
\newblock In \emph{Proceedings of the IEEE/CVF international conference on computer vision}, pages 9298--9309, 2023{\natexlab{a}}.

\bibitem[Liu et~al.(2023{\natexlab{b}})Liu, Lin, Zeng, Long, Liu, Komura, and Wang]{liu2023syncdreamer}
Yuan Liu, Cheng Lin, Zijiao Zeng, Xiaoxiao Long, Lingjie Liu, Taku Komura, and Wenping Wang.
\newblock Syncdreamer: Generating multiview-consistent images from a single-view image.
\newblock \emph{arXiv preprint arXiv:2309.03453}, 2023{\natexlab{b}}.

\bibitem[Long et~al.(2024)Long, Guo, Lin, Liu, Dou, Liu, Ma, Zhang, Habermann, Theobalt, et~al.]{long2024wonder3d}
Xiaoxiao Long, Yuan-Chen Guo, Cheng Lin, Yuan Liu, Zhiyang Dou, Lingjie Liu, Yuexin Ma, Song-Hai Zhang, Marc Habermann, Christian Theobalt, et~al.
\newblock Wonder3d: Single image to 3d using cross-domain diffusion.
\newblock In \emph{Proceedings of the IEEE/CVF Conference on Computer Vision and Pattern Recognition}, pages 9970--9980, 2024.

\bibitem[Miao et~al.(2024)Miao, Luo, and Yang]{miao2024pla4d}
Qiaowei Miao, Yawei Luo, and Yi Yang.
\newblock Pla4d: Pixel-level alignments for text-to-4d gaussian splatting.
\newblock \emph{arXiv preprint arXiv:2405.19957}, 2024.

\bibitem[Mildenhall et~al.(2021)Mildenhall, Srinivasan, Tancik, Barron, Ramamoorthi, and Ng]{mildenhall2021nerf}
Ben Mildenhall, Pratul~P Srinivasan, Matthew Tancik, Jonathan~T Barron, Ravi Ramamoorthi, and Ren Ng.
\newblock Nerf: Representing scenes as neural radiance fields for view synthesis.
\newblock \emph{Communications of the ACM}, 65\penalty0 (1):\penalty0 99--106, 2021.

\bibitem[Mou et~al.(2024)Mou, Wang, Xie, Wu, Zhang, Qi, and Shan]{mou2024t2i}
Chong Mou, Xintao Wang, Liangbin Xie, Yanze Wu, Jian Zhang, Zhongang Qi, and Ying Shan.
\newblock T2i-adapter: Learning adapters to dig out more controllable ability for text-to-image diffusion models.
\newblock In \emph{Proceedings of the AAAI Conference on Artificial Intelligence}, pages 4296--4304, 2024.

\bibitem[Niemeyer et~al.(2022)Niemeyer, Barron, Mildenhall, Sajjadi, Geiger, and Radwan]{niemeyer2022regnerf}
Michael Niemeyer, Jonathan~T Barron, Ben Mildenhall, Mehdi~SM Sajjadi, Andreas Geiger, and Noha Radwan.
\newblock Regnerf: Regularizing neural radiance fields for view synthesis from sparse inputs.
\newblock In \emph{Proceedings of the IEEE/CVF Conference on Computer Vision and Pattern Recognition}, pages 5480--5490, 2022.

\bibitem[Podell et~al.(2023)Podell, English, Lacey, Blattmann, Dockhorn, M{\"u}ller, Penna, and Rombach]{podell2023sdxl}
Dustin Podell, Zion English, Kyle Lacey, Andreas Blattmann, Tim Dockhorn, Jonas M{\"u}ller, Joe Penna, and Robin Rombach.
\newblock Sdxl: Improving latent diffusion models for high-resolution image synthesis.
\newblock \emph{arXiv preprint arXiv:2307.01952}, 2023.

\bibitem[Poole et~al.(2022)Poole, Jain, Barron, and Mildenhall]{poole2022dreamfusion}
Ben Poole, Ajay Jain, Jonathan~T Barron, and Ben Mildenhall.
\newblock Dreamfusion: Text-to-3d using 2d diffusion.
\newblock \emph{arXiv preprint arXiv:2209.14988}, 2022.

\bibitem[Pumarola et~al.(2021)Pumarola, Corona, Pons-Moll, and Moreno-Noguer]{pumarola2021d}
Albert Pumarola, Enric Corona, Gerard Pons-Moll, and Francesc Moreno-Noguer.
\newblock D-nerf: Neural radiance fields for dynamic scenes.
\newblock In \emph{Proceedings of the IEEE/CVF Conference on Computer Vision and Pattern Recognition}, pages 10318--10327, 2021.

\bibitem[Radford et~al.(2021)Radford, Kim, Hallacy, Ramesh, Goh, Agarwal, Sastry, Askell, Mishkin, Clark, et~al.]{radford2021learning}
Alec Radford, Jong~Wook Kim, Chris Hallacy, Aditya Ramesh, Gabriel Goh, Sandhini Agarwal, Girish Sastry, Amanda Askell, Pamela Mishkin, Jack Clark, et~al.
\newblock Learning transferable visual models from natural language supervision.
\newblock In \emph{International conference on machine learning}, pages 8748--8763. PMLR, 2021.

\bibitem[Ren et~al.(2023)Ren, Pan, Tang, Zhang, Cao, Zeng, and Liu]{ren2023dreamgaussian4d}
Jiawei Ren, Liang Pan, Jiaxiang Tang, Chi Zhang, Ang Cao, Gang Zeng, and Ziwei Liu.
\newblock Dreamgaussian4d: Generative 4d gaussian splatting.
\newblock \emph{arXiv preprint arXiv:2312.17142}, 2023.

\bibitem[Ren et~al.(2024)Ren, Xie, Mirzaei, Liang, Zeng, Kreis, Liu, Torralba, Fidler, Kim, et~al.]{ren2024l4gm}
Jiawei Ren, Kevin Xie, Ashkan Mirzaei, Hanxue Liang, Xiaohui Zeng, Karsten Kreis, Ziwei Liu, Antonio Torralba, Sanja Fidler, Seung~Wook Kim, et~al.
\newblock L4gm: Large 4d gaussian reconstruction model.
\newblock \emph{arXiv preprint arXiv:2406.10324}, 2024.

\bibitem[Rombach et~al.(2022)Rombach, Blattmann, Lorenz, Esser, and Ommer]{rombach2022high}
Robin Rombach, Andreas Blattmann, Dominik Lorenz, Patrick Esser, and Bj{\"o}rn Ommer.
\newblock High-resolution image synthesis with latent diffusion models.
\newblock In \emph{Proceedings of the IEEE/CVF conference on computer vision and pattern recognition}, pages 10684--10695, 2022.

\bibitem[Shi et~al.(2023)Shi, Wang, Ye, Long, Li, and Yang]{shi2023mvdream}
Yichun Shi, Peng Wang, Jianglong Ye, Mai Long, Kejie Li, and Xiao Yang.
\newblock Mvdream: Multi-view diffusion for 3d generation.
\newblock \emph{arXiv preprint arXiv:2308.16512}, 2023.

\bibitem[Smart et~al.(2024)Smart, Zheng, Laina, and Prisacariu]{smart2024splatt3r}
Brandon Smart, Chuanxia Zheng, Iro Laina, and Victor~Adrian Prisacariu.
\newblock Splatt3r: Zero-shot gaussian splatting from uncalibarated image pairs.
\newblock \emph{arXiv preprint arXiv:2408.13912}, 2024.

\bibitem[Sun et~al.(2024)Sun, Guo, Wan, Yan, Yin, Zhou, Liao, and Li]{sun2024eg4d}
Qi Sun, Zhiyang Guo, Ziyu Wan, Jing~Nathan Yan, Shengming Yin, Wengang Zhou, Jing Liao, and Houqiang Li.
\newblock Eg4d: Explicit generation of 4d object without score distillation.
\newblock \emph{arXiv preprint arXiv:2405.18132}, 2024.

\bibitem[Tang et~al.(2023)Tang, Ren, Zhou, Liu, and Zeng]{tang2023dreamgaussian}
Jiaxiang Tang, Jiawei Ren, Hang Zhou, Ziwei Liu, and Gang Zeng.
\newblock Dreamgaussian: Generative gaussian splatting for efficient 3d content creation.
\newblock \emph{arXiv preprint arXiv:2309.16653}, 2023.

\bibitem[Tang et~al.(2024)Tang, Chen, Chen, Wang, Zeng, and Liu]{tang2024lgm}
Jiaxiang Tang, Zhaoxi Chen, Xiaokang Chen, Tengfei Wang, Gang Zeng, and Ziwei Liu.
\newblock Lgm: Large multi-view gaussian model for high-resolution 3d content creation.
\newblock In \emph{European Conference on Computer Vision}, pages 1--18. Springer, 2024.

\bibitem[Unterthiner et~al.(2018)Unterthiner, Van~Steenkiste, Kurach, Marinier, Michalski, and Gelly]{unterthiner2018towards}
Thomas Unterthiner, Sjoerd Van~Steenkiste, Karol Kurach, Raphael Marinier, Marcin Michalski, and Sylvain Gelly.
\newblock Towards accurate generative models of video: A new metric \& challenges.
\newblock \emph{arXiv preprint arXiv:1812.01717}, 2018.

\bibitem[Villegas et~al.(2022)Villegas, Babaeizadeh, Kindermans, Moraldo, Zhang, Saffar, Castro, Kunze, and Erhan]{villegas2022phenaki}
Ruben Villegas, Mohammad Babaeizadeh, Pieter-Jan Kindermans, Hernan Moraldo, Han Zhang, Mohammad~Taghi Saffar, Santiago Castro, Julius Kunze, and Dumitru Erhan.
\newblock Phenaki: Variable length video generation from open domain textual descriptions.
\newblock In \emph{International Conference on Learning Representations}, 2022.

\bibitem[Wang et~al.(2004)Wang, Bovik, Sheikh, and Simoncelli]{wang2004image}
Zhou Wang, Alan~C Bovik, Hamid~R Sheikh, and Eero~P Simoncelli.
\newblock Image quality assessment: from error visibility to structural similarity.
\newblock \emph{IEEE transactions on image processing}, 13\penalty0 (4):\penalty0 600--612, 2004.

\bibitem[Wang et~al.(2024)Wang, Lu, Wang, Bao, Li, Su, and Zhu]{wang2024prolificdreamer}
Zhengyi Wang, Cheng Lu, Yikai Wang, Fan Bao, Chongxuan Li, Hang Su, and Jun Zhu.
\newblock Prolificdreamer: High-fidelity and diverse text-to-3d generation with variational score distillation.
\newblock \emph{Advances in Neural Information Processing Systems}, 36, 2024.

\bibitem[Wu et~al.(2024)Wu, Yi, Fang, Xie, Zhang, Wei, Liu, Tian, and Wang]{wu20244d}
Guanjun Wu, Taoran Yi, Jiemin Fang, Lingxi Xie, Xiaopeng Zhang, Wei Wei, Wenyu Liu, Qi Tian, and Xinggang Wang.
\newblock 4d gaussian splatting for real-time dynamic scene rendering.
\newblock In \emph{Proceedings of the IEEE/CVF Conference on Computer Vision and Pattern Recognition}, pages 20310--20320, 2024.

\bibitem[Wu et~al.(2023)Wu, Ge, Wang, Lei, Gu, Shi, Hsu, Shan, Qie, and Shou]{wu2023tune}
Jay~Zhangjie Wu, Yixiao Ge, Xintao Wang, Stan~Weixian Lei, Yuchao Gu, Yufei Shi, Wynne Hsu, Ying Shan, Xiaohu Qie, and Mike~Zheng Shou.
\newblock Tune-a-video: One-shot tuning of image diffusion models for text-to-video generation.
\newblock In \emph{Proceedings of the IEEE/CVF International Conference on Computer Vision}, pages 7623--7633, 2023.

\bibitem[Xie et~al.(2024)Xie, Yao, Voleti, Jiang, and Jampani]{xie2024sv4d}
Yiming Xie, Chun-Han Yao, Vikram Voleti, Huaizu Jiang, and Varun Jampani.
\newblock Sv4d: Dynamic 3d content generation with multi-frame and multi-view consistency.
\newblock \emph{arXiv preprint arXiv:2407.17470}, 2024.

\bibitem[Xu et~al.(2024)Xu, Nie, Liu, Liu, Kautz, Wang, and Vahdat]{xu2024camco}
Dejia Xu, Weili Nie, Chao Liu, Sifei Liu, Jan Kautz, Zhangyang Wang, and Arash Vahdat.
\newblock Camco: Camera-controllable 3d-consistent image-to-video generation.
\newblock \emph{arXiv preprint arXiv:2406.02509}, 2024.

\bibitem[Yang et~al.(2024{\natexlab{a}})Yang, Gao, Zhou, Jiao, Zhang, and Jin]{yang2024deformable}
Ziyi Yang, Xinyu Gao, Wen Zhou, Shaohui Jiao, Yuqing Zhang, and Xiaogang Jin.
\newblock Deformable 3d gaussians for high-fidelity monocular dynamic scene reconstruction.
\newblock In \emph{Proceedings of the IEEE/CVF Conference on Computer Vision and Pattern Recognition}, pages 20331--20341, 2024{\natexlab{a}}.

\bibitem[Yang et~al.(2024{\natexlab{b}})Yang, Pan, Gu, and Zhang]{yang2024diffusion}
Zeyu Yang, Zijie Pan, Chun Gu, and Li Zhang.
\newblock Diffusion $^{2}$: Dynamic 3d content generation via score composition of orthogonal diffusion models.
\newblock \emph{arXiv preprint arXiv:2404.02148}, 2024{\natexlab{b}}.

\bibitem[Yi et~al.(2023)Yi, Fang, Wu, Xie, Zhang, Liu, Tian, and Wang]{yi2023gaussiandreamer}
Taoran Yi, Jiemin Fang, Guanjun Wu, Lingxi Xie, Xiaopeng Zhang, Wenyu Liu, Qi Tian, and Xinggang Wang.
\newblock Gaussiandreamer: Fast generation from text to 3d gaussian splatting with point cloud priors.
\newblock \emph{arXiv preprint arXiv:2310.08529}, 2023.

\bibitem[Yu et~al.(2024)Yu, Xing, Yuan, Hu, Li, Huang, Gao, Wong, Shan, and Tian]{yu2024viewcrafter}
Wangbo Yu, Jinbo Xing, Li Yuan, Wenbo Hu, Xiaoyu Li, Zhipeng Huang, Xiangjun Gao, Tien-Tsin Wong, Ying Shan, and Yonghong Tian.
\newblock Viewcrafter: Taming video diffusion models for high-fidelity novel view synthesis.
\newblock \emph{arXiv preprint arXiv:2409.02048}, 2024.

\bibitem[Yuan et~al.(2024)Yuan, Kobbelt, Liu, Zhang, Wan, Lai, and Gao]{yuan20244dynamic}
Yu-Jie Yuan, Leif Kobbelt, Jiwen Liu, Yuan Zhang, Pengfei Wan, Yu-Kun Lai, and Lin Gao.
\newblock 4dynamic: Text-to-4d generation with hybrid priors.
\newblock \emph{arXiv preprint arXiv:2407.12684}, 2024.

\bibitem[Zeng et~al.(2025)Zeng, Jiang, Zhu, Lu, Lin, Zhu, Hu, Cao, and Yao]{zeng2025stag4d}
Yifei Zeng, Yanqin Jiang, Siyu Zhu, Yuanxun Lu, Youtian Lin, Hao Zhu, Weiming Hu, Xun Cao, and Yao Yao.
\newblock Stag4d: Spatial-temporal anchored generative 4d gaussians.
\newblock In \emph{European Conference on Computer Vision}, pages 163--179. Springer, 2025.

\bibitem[Zhang et~al.(2024{\natexlab{a}})Zhang, Wu, Liu, Zhao, Ran, Gu, Gao, and Shou]{zhang2024show}
David~Junhao Zhang, Jay~Zhangjie Wu, Jia-Wei Liu, Rui Zhao, Lingmin Ran, Yuchao Gu, Difei Gao, and Mike~Zheng Shou.
\newblock Show-1: Marrying pixel and latent diffusion models for text-to-video generation.
\newblock \emph{International Journal of Computer Vision}, pages 1--15, 2024{\natexlab{a}}.

\bibitem[Zhang et~al.(2024{\natexlab{b}})Zhang, Chen, Wang, Liu, Wang, and Qiao]{zhang20244diffusion}
Haiyu Zhang, Xinyuan Chen, Yaohui Wang, Xihui Liu, Yunhong Wang, and Yu Qiao.
\newblock 4diffusion: Multi-view video diffusion model for 4d generation.
\newblock \emph{arXiv preprint arXiv:2405.20674}, 2024{\natexlab{b}}.

\bibitem[Zhang et~al.(2023)Zhang, Rao, and Agrawala]{zhang2023adding}
Lvmin Zhang, Anyi Rao, and Maneesh Agrawala.
\newblock Adding conditional control to text-to-image diffusion models.
\newblock In \emph{Proceedings of the IEEE/CVF International Conference on Computer Vision}, pages 3836--3847, 2023.

\bibitem[Zhang et~al.(2018)Zhang, Isola, Efros, Shechtman, and Wang]{zhang2018unreasonable}
Richard Zhang, Phillip Isola, Alexei~A Efros, Eli Shechtman, and Oliver Wang.
\newblock The unreasonable effectiveness of deep features as a perceptual metric.
\newblock In \emph{Proceedings of the IEEE conference on computer vision and pattern recognition}, pages 586--595, 2018.

\bibitem[Zhao et~al.(2023)Zhao, Yan, Xie, Hong, Li, and Lee]{zhao2023animate124}
Yuyang Zhao, Zhiwen Yan, Enze Xie, Lanqing Hong, Zhenguo Li, and Gim~Hee Lee.
\newblock Animate124: Animating one image to 4d dynamic scene.
\newblock \emph{arXiv preprint arXiv:2311.14603}, 2023.

\bibitem[Zhu et~al.(2024)Zhu, He, Tang, Guo, Chen, and Bian]{zhu2024compositional}
Hanxin Zhu, Tianyu He, Anni Tang, Junliang Guo, Zhibo Chen, and Jiang Bian.
\newblock Compositional 3d-aware video generation with llm director.
\newblock \emph{arXiv preprint arXiv:2409.00558}, 2024.

\end{thebibliography}
